\pdfoutput=1

\documentclass[11pt]{article}

\usepackage[preprint]{acl}

\usepackage{times}
\usepackage{latexsym}

\usepackage[T1]{fontenc}
\usepackage{adjustbox}
\usepackage[utf8]{inputenc}

\usepackage{microtype}

\usepackage{inconsolata}

\usepackage{graphicx}

\usepackage{array}
\usepackage{makecell}
\usepackage{multirow}
\usepackage{booktabs}
\usepackage{tcolorbox} 
\usepackage{listings} 
\usepackage{xcolor} 
\usepackage{subcaption}
\usepackage{amssymb}
\usepackage{pifont}
\usepackage{tabularx}
\tcbuselibrary{listings, breakable, skins}

\newcommand{\cmark}{\ding{51}} 
\title{EduBench: A Comprehensive Benchmarking Dataset for Evaluating Large Language Models in Diverse Educational Scenarios}

\author{
Bin Xu$^*$, Yu Bai$^*$, Huashan Sun$^*$, Yiguan Lin\thanks{Equal contribution.} \\
\textbf{Siming Liu, Xinyue Liang, Yaolin Li} \\
\textbf{Zhuangzhi Dong, Jingren Zhang, Yufan Deng, Xinyu Zou} \hspace{1mm} \\
\textbf{Yang Gao\thanks{Corresponding author.}} 
\textbf{, Heyan Huang}\\
School of Computer Science and Technology, \\ Beijing Institute of Technology, Beijing, China \\
\{binxu, yubai, gyang\}@bit.edu.cn
}

\usepackage{tcolorbox}
\usepackage{xcolor}
\usepackage{longtable}
\usepackage{amsmath}

\begin{document}
\maketitle
\begin{abstract}
As large language models continue to advance, their application in educational contexts remains underexplored and under-optimized. In this paper, we address this gap by introducing the first diverse benchmark tailored for educational scenarios, incorporating synthetic data containing 9 major scenarios and over 4,000 distinct educational contexts.
To enable comprehensive assessment, we propose a set of multi-dimensional evaluation metrics that cover 12 critical aspects relevant to both teachers and students. We further apply human annotation to ensure the effectiveness of the model-generated evaluation responses. Additionally, we succeed to train a relatively small-scale model on our constructed dataset and demonstrate that it can achieve performance comparable to state-of-the-art large models (e.g., Deepseek V3, Qwen Max) on the test set. Overall, this work provides a practical foundation for the development and evaluation of education-oriented language models. 


\end{abstract}

\begin{figure*}[t]
  \centering
  \includegraphics[width=\linewidth]{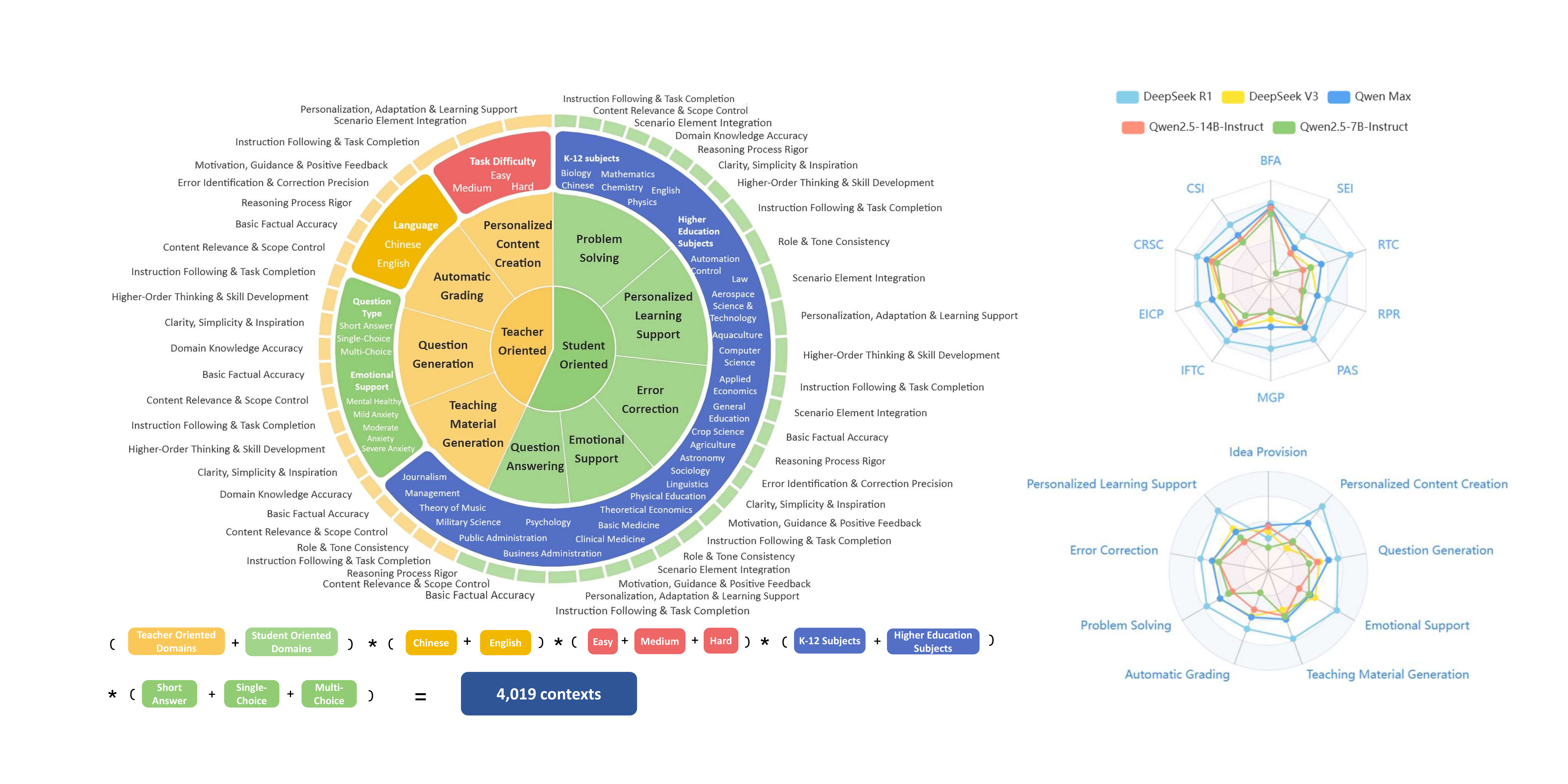} \hfill
  \caption {The left section presents our 9 educational scenarios, along with their multi-dimensional educational contexts and corresponding metrics. The right section illustrates the results from human evaluation on EduBench.}
\end{figure*}

\section{Introduction}

Large Language Models (LLMs) have recently shown remarkable potential in educational contexts, offering capabilities such as problem-solving, interactive dialogue, and decision-making~\citep{jiaEQGRACEExaminationTypeQuestion2021, rouzegarGenerativeAIEnhancing2024,li2025fundamental}. 
These abilities make LLMs promising tools for tasks ranging from personalized tutoring to educational content generation. However, despite growing interest, research on their practical deployment in education remains limited. 

A key limitation of previous work is their narrow focus on knowledge-intensive tasks~\citep{rein_gpqa_2023,team_supergpqa_2025,cobbe2021trainingverifierssolvemath}, which fails to reflect the diverse educational scenarios encountered in real-world settings. 
These efforts often overlook the complexity introduced by varying roles (e.g., professors, students, psychological counselors), whose needs, goals, and interaction styles differ significantly. 
More importantly, existing benchmarks~\citep{huang_edunlp_2024, koutcheme_benchmarking_2024, yang_social_2024} rarely align \textbf{task interactions with learners' cognitive levels and scenario-specific objectives}, leading to evaluation dimensions that lack pedagogical relevance. 
In contrast, our work introduces \textbf{EduBench}, a benchmark designed not only to support educational applications, but also to promote the development of robust and goal-aligned evaluation mechanisms that reflect the diversity of modern educational needs. 
By going beyond purely knowledge-based tasks~\citep{koutcheme_using_2024,ng_educational_2024,huangGeneratingEducationalMaterials2024a}, the EduBench re-centers educational AI research on the core values of education—holistic development, personalized support, and context-aware learning—while systematically constructing data that captures the rich landscape of roles, domains, and learning scenarios.

For constructing data, we create a diverse dataset considering 9 different educational scenarios, including assignment judging, proposing a study plan given the profile of specific students, providing psychological health. 
We design several educational contexts in each scenario to further facilitate the diversity of the data, such as question difficulties, student grades (e.g., elementary school students, high school students, graduate students, etc.), and different subjects. Collectively, these categories cover over 4,000 diverse educational contexts. 
For these different contexts, we create different querying data,
resulting in a dataset containing 18,821 data points. 
To ensure the quality of the data, we proportionally sample a total of 500 queries across different scenarios and evaluate them using both human raters and multiple LLMs. All evaluations follow a set of carefully designed human-written criteria.


In designing our evaluation metrics, we focus on three principal aspects. 
First, \textbf{Scenario Adaptation} determines if the model’s response appropriately addresses the conditions and constraints defined by specific educational scenarios, task instructions, and role-playing requirements. 
Second, \textbf{Factual \& Reasoning Accuracy} scrutinizes both the factual correctness of the information presented and the logical soundness of the reasoning applied. 
Third, \textbf{Pedagogical Application} examines the response’s adherence to established educational principles and its potential to positively impact students’ learning experiences. Each major aspect contains 4 distinct sub-metrics to evaluate the responses in a finer-grained way. 
These metrics are systematically allocated across nine educational scenarios to ensure comprehensive coverage of diverse task demands.

When evaluating the responses of different models, we notice a trade-off between accuracy and cost when comparing human and LLM evaluators. Hence, we 
choose to investigate the evaluation capabilities of different LLMs by comparing their evaluative responses with those of human judges in a small test set consisting of 198 diverse data points.
Our experiments on several LLMs reveal that DeepSeek V3 achieves the best alignment with human annotators. Results of using it to benchmark five different LLMs show that: (1) The models’ understanding of the scoring guidelines remains imperfect. (2) Not all models, when used as evaluators, align closely with human ratings; DeepSeek V3 stands out for its consistency, whereas GPT-4o performs relatively weaker. (3) Smaller models in general would perform worse than large models. 

As smaller language models typically underperform significantly larger models, we further highlight the utility of our data in boosting the performance of smaller models. This is achieved through knowledge distillation from powerful larger models, aiming to bridge the performance gap. We implement a multi-source distillation strategy, extracting expertise for each scenario from the LLMs with the highest performance on that scenario. Our results demonstrate that a 7B model can attain performance comparable to the state-of-the-art 671B DeepSeek V3 model using a dataset size of around 17,000 training samples. 
We summarize our contribution as follows:
\begin{enumerate}
    \item We present the first LLM-powered educational benchmark with the largest scalable scenario and context collection (4,000+ contexts), along with a 12-dimension evaluation system.
    \item We establish a series of findings and will release all the model-generated and human-annotated data that could benefit the LLM research community regarding both educational applications and LLM-based evaluations.
    \item We show that our data could benefit smaller models to achieve comparable performance with powerful state-of-the-art LLMs.
\end{enumerate}

\begin{figure}[t]
  \centering
  \includegraphics[width=\linewidth]{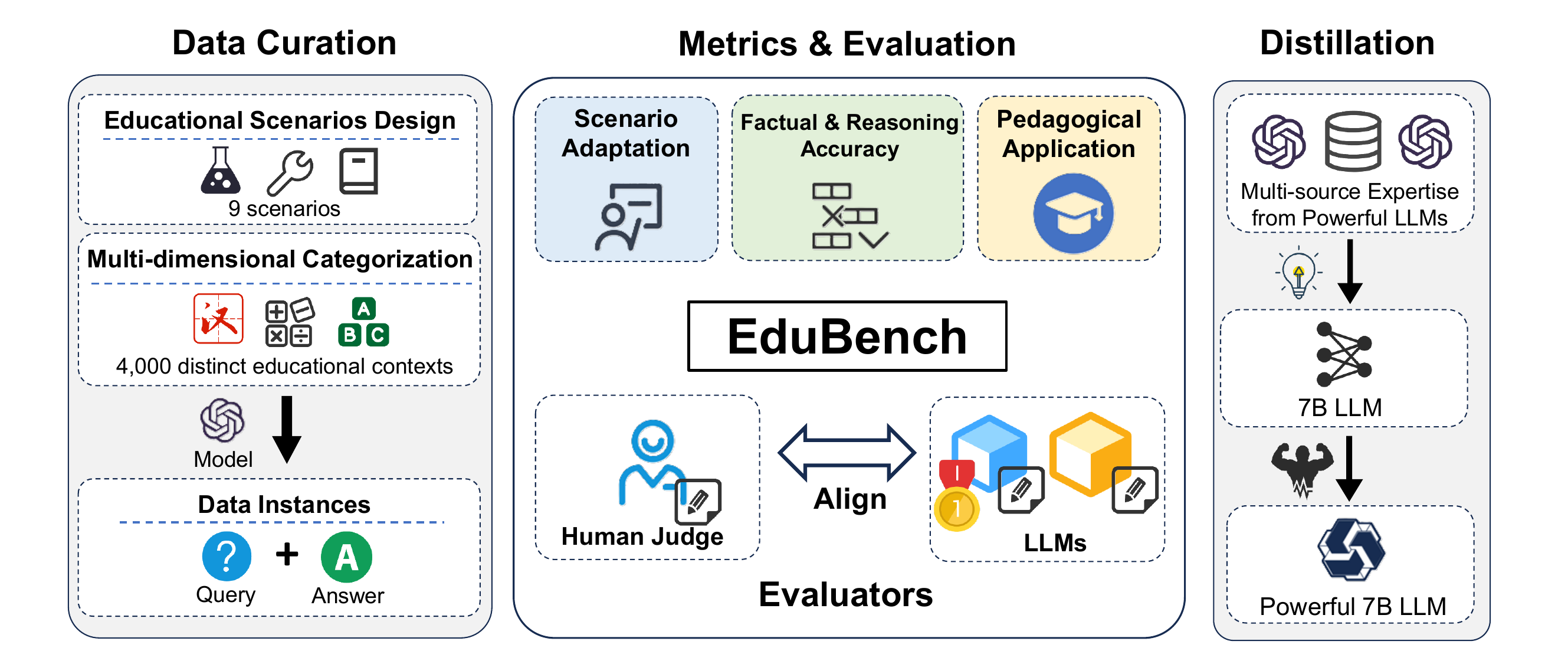} \hfill
  \caption {Overview of EduBench, with data curation on the left, evaluation principles and human–LLM alignment in the middle, and downstream performance gains for smaller models on the right.}
\end{figure}

\section{Dataset Collection}
In this section, we describe the dataset construction methodology used in this paper. Our objective is to build a comprehensive benchmark that reflects realistic and diverse scenarios in the educational domain, covering both students’ and teachers’ needs across multiple functional capacities with varied levels of difficulty and content modalities. 

To achieve this goal, we choose to synthesize data using LLMs due to the limitations of real-world educational data. In practice, data from certain domains remain sparse and exhibit unstable quality, such as student psychological counseling. To address this issue, we generate data based on human-written seed scenarios, which provide structured guidance for LLMs and help mitigate data sparsity in underrepresented scenarios.


\subsection{Scenario Design}
\label{sec:domain_design}
We categorize and organize a set of key education-related tasks~\citep{gan_large_2023,wang_large_2024}, each reflecting a distinct dimension of education. 
Based on these tasks, we define a “scenario” as a representative educational scenario that involves typical user roles, cognitive skill demands, input-output formats, and evaluation criteria. 
Building on the distinction in typical user roles, we categorize these tasks into Student-Oriented Scenarios~\citep{ng_educational_2024,team_supergpqa_2025,koutcheme_using_2024} and Teacher-Oriented Scenarios~\citep{zhang_simulating_2025,ghanemQuestionGenerationReading2022a,huangGeneratingEducationalMaterials2024a}.
This conceptualization enables us to structure a wide range of tasks grounded in real-world educational practice.

To support both learner assistance and instructional tasks, we categorize domains into two primary types.

\textbf{Student-Oriented Scenarios:} Problem Solving (Q\&A),  Error Correction (EC), Idea Provision (IP), Personalized Learning Support (PLS), Emotional Support (ES).

\textbf{Teacher-Oriented Scenarios:}
 Question Generation (QG), Automatic Grading (AG), Teaching Material Generation (TMG), Personalized Content Creation (PCC).
    

These scenarios simulate key educational roles and scenarios, enabling realistic evaluation of LLM capabilities in classroom, tutoring, and curriculum design settings. A detailed breakdown of each scenario is provided in Appendix~\ref{appendix:domain-design}.
\subsection{Educational Context Design}

To ensure scenario realism, learner alignment, and meaningful evaluation, we construct diverse educational domain contexts that reflect the conditions under which tasks naturally occur. Each context is designed to capture variation across four primary dimensions: \textbf{subject taxonomy}, \textbf{task difficulty}, \textbf{language setting}, and \textbf{question type}. This diversity enables EduBench to cover a wide range of real-world educational scenarios, from K–12 to postgraduate levels, and from basic recall to complex reasoning.

Following the disciplinary category design of EduNLP~\citep{huang_edunlp_2024} and  SuperGPQA~\citep{team_supergpqa_2025}, we categorize tasks based on educational stages (e.g., K–12 vs. higher education) and align them with different cognitive and pedagogical goals. Each task is assigned a difficulty level (easy, medium, hard) to reflect expected learner proficiency. EduBench currently supports both \textbf{Chinese and English} tasks to encourage multilingual model evaluation. Question types are specified based on domain functionality: most scenarios adopt common types such as \textit{Single Choice}, \textit{Multiple Choice}, and \textit{Short Answer}, while specialized scenarios follow scenario-specific definitions. 
Details of all the designs of the educational context can be found in Appendix~\ref{appendix:educational-context}. 

\subsection{Question Generation for Scenario Tasks}
\label{sec:question-gen}



Building on the structured scenario design, we generate benchmark data using a systematic and scalable pipeline tailored to educational scenarios. First, we organize our coverage across nine major scenarios grounded in an educational competency framework (see Appendix~\ref{appendix:domain-design}). Within each scenario, we apply multi-dimensional categorization, including subject taxonomy, task difficulty, and question type, to define fine-grained task scenarios.

For each scenario, we design prompt templates that reflect realistic user intents, then use GPT-4o to generate consistent data instances.
For example, in the mathematics domain, a scenario like “Middle School – Short Answer Question” guides the model in producing relevant QA pairs. 
Further prompt design details and generated samples are available in Appendix~\ref{appendix:prompt-examples}.

\subsection{Data Quality Assessment}

\begin{figure}[h]
    \centering
    \includegraphics[width=\linewidth]{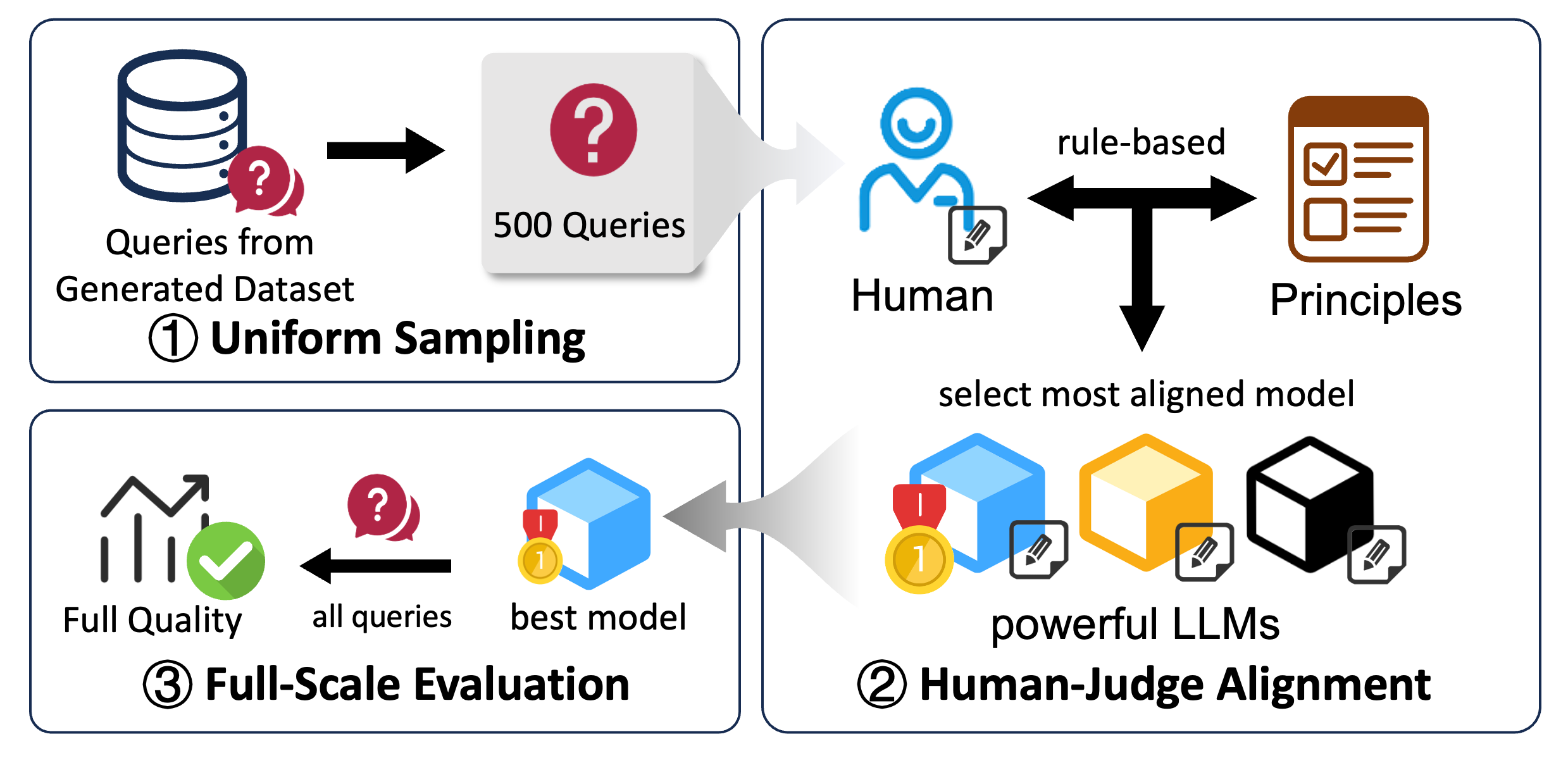} 
    \caption {Workflow of Selecting the Most Human-Aligned LLM and Conducting Full-Scale Evaluation.}
    \label{fig:query-eval}
\end{figure}

To verify the quality of queries generated by LLMs, as in Figure~\ref{fig:query-eval}, we adapt a human-guided hybrid evaluation mechanism. Guided by educational usability requirements, we manually design a set of scenario-aligned evaluation criteria, consisting of two universal dimensions — linguistic quality and content quality — together with task-specific dimensions corresponding to each scenario (see Appendix~\ref{query-evaluation-principles}). Based on these criteria, we proportionally sample approximately 500 queries across scenarios for dual evaluation: (1) human annotators score the queries, and (2) several high-performing models generate automatic scores under the same rules. We use the Mean Absolute Error (MAE) to measure how much a model’s evaluation of a query deviates from human ratings. For each evaluation dimension, we compute the absolute difference between the model score and the human score, and then average this value over all queries that contain this dimension. This yields the model’s deviation from human judgments for that specific dimension.

By comparing the consistency between model and human ratings in Table~\ref{tab:query_evaluation}, we select QwQ-32B, the model most aligned with human judgments, as the final automatic evaluator, achieving a balance between cost efficiency and assessment reliability. This model is then applied to conduct full-scale quality evaluation over the dataset. Results in Table~\ref{tab:qwq_query_evaluation} demonstrate that the dataset maintains high overall quality. The generated queries consistently exhibit strong performance in linguistic quality, content reliability, and task appropriateness, confirming the effectiveness of our design.

\begin{table}[htbp]
    \centering
    \caption{Query evaluation: the subset refers to queries randomly selected for human and model evaluation and the full refers to the entire benchmark.}
    \label{tab:qwq_query_evaluation}
    \begin{adjustbox}{width=\linewidth, keepaspectratio}
        \begin{tabular}{l c c c}
            \toprule
            \textbf{Task Name} & \textbf{Human-subset} & \textbf{QwQ-subset} & \textbf{QwQ-full} \\
            \midrule
            Error Correction & 8.26 & 9.38 & 8.39 \\
            Problem Solving / Idea Provision & 9.07 & 9.11 & 9.01 \\
            Automatic Grading & 8.31 & 8.50 & 8.72 \\
            Teaching Material Generation & 6.98 & 8.81 & 8.78 \\
            Question Generation & 8.81 & 9.15 & 8.89 \\
            Emotional Support & 8.07 & 8.77 & 8.88 \\
            Personalized Content Creation & 8.17 & 8.78 & 8.95 \\
            Personalized Learning Support & 8.31 & 9.00 & 9.00 \\
            \bottomrule
        \end{tabular}
    \end{adjustbox}
\end{table}

\section{Evaluation Metric Design}
\label{sec:metric-design}
With a set of well-established query data, we proceed to assess the ability of frontier LLMs to properly fulfill the diverse requirements (reflected by the queries) and generate high-quality responses. 

In this section, we design a set of evaluation metrics from different dimensions to achieve such a goal. Evaluating complex domains such as education is challenging due to multiple participants and multidimensionality.
Previous studies have employed large language models (LLMs) for evaluation purposes~\citep{koutcheme_benchmarking_2024,team_supergpqa_2025, yang_social_2024,wang2024patient,ng_educational_2024}. 
Building on this foundation, we proposed a newly designed, structured set of evaluation metrics that include 12 dimensions and reflect both pedagogical goals and scenario-specific expectations to enhance the accuracy and interpretability of model evaluation. Our design comprises three core dimensions: Scenario Adaptation, Factual \& Reasoning Accuracy, Pedagogical Application.


\subsection{Overall Design}
To enhance the accuracy and interpretability of model evaluation, we design a series of evaluation metrics. By incorporating these criteria into the evaluation prompts, we guide the model to assign scores according to the specified metrics based on the given data, while also producing detailed justifications during the evaluation process. The approach of introducing metrics or principles in evaluation has been widely validated~\citep{liu2025inference, sharma2025constitutional, bai2022constitutional}, with some being manually designed and others generated by the model before or during the evaluation. 

\subsection{Metric Design}
We design distinct evaluation metrics based on 3 scenarios, with 4 metrics for each scenario to cover its key aspects, resulting in 12 different metrics.

\paragraph{Scenario Adaptation} The Scenario Adaptation metric evaluates whether model responses are contextually appropriate and aligned with educational expectations. It is assessed across four dimensions: 1) \textit{Instruction Following \& Task Completion}, 2) \textit{Role \& Tone Consistency}, 3) \textit{Content Relevance \& Scope Control}, 4) \textit{Scenario Element Integration}.

\paragraph{Factual \& Reasoning Accuracy} The Factual \& Reasoning Accuracy metric assesses the correctness of information and the rigor of reasoning in model responses. It includes four sub-metrics: 1) \textit{Basic Factual Accuracy}, 2) \textit{Domain Knowledge Accuracy}, 3) \textit{Reasoning Process Rigor}, 4) \textit{Error Identification \& Correction Precision}.

\paragraph{Pedagogical Application} This metric evaluates whether responses embody sound educational principles and effectively support student learning. It consists of the following sub-metrics: 1) \textit{Clarity, Simplicity \& Inspiration}, 2) \textit{Motivation, Guidance \& Positive Feedback}, 3) \textit{Personalization, Adaptation \& Learning Support}, 4) \textit{Higher-Order Thinking \& Skill Development}.

The detailed explanations of these metrics are provided in Appendix~\ref{appendix:evaluation_metric_design}.

\subsection{Dynamic Metric Allocation}
\label{sec:metric_allocation}
Given the diversity of educational tasks covered in EduBench, a one-size-fits-all evaluation approach is insufficient. Not all metrics are equally relevant or applicable across the nine distinct scenarios. For example, Emotional Support tasks emphasize contextual empathy and scenario alignment more than factual precision, while scenarios like Problem Solving and Question Generation rely heavily on rigorous reasoning and factual correctness.

To address this, we design a flexible evaluation framework that dynamically allocates appropriate metrics based on the specific requirements of each scenario. Each scenario is associated with a tailored subset of evaluation dimensions that best reflect its instructional goals, content characteristics, and target outcomes. This ensures both fairness and relevance in the evaluation process. Detailed metric-scenario mappings and allocation rules are provided in Appendix~\ref{appendix:metric-allocation}.

\subsection{Human-guided LLM-based Evaluation}
\label{sec:human_guided}
Evaluating open-ended educational tasks is challenging due to the subjectivity and complexity of responses. While human annotation offers high-quality judgments, it is costly and difficult to scale. On the other hand, relying solely on LLM-based evaluation raises concerns about reliability and consistency across diverse scenarios. 

We adopt a human-guided evaluation framework by constructing a high-quality test set of 198 diverse examples (11 per scenario in both English and Chinese), annotated by expert judges across various task types, to measure the alignment between LLMs and human judgments. 

This approach enables scalable and efficient assessment aligned with human standards.

\begin{table*}[t]
\centering

\begin{subtable}[t]{0.68\linewidth}
\centering
\begin{adjustbox}{width=\linewidth}
\begin{tabular}{l|l|cccccccccccc|c}
\toprule
\textbf{Evaluator} & \textbf{Model} & \textbf{BFA} & \textbf{CSI} & \textbf{CRSC} & \textbf{DKA} & \textbf{EICP} & \textbf{HOTS} & \textbf{IFTC} & \textbf{MGP} & \textbf{PAS} & \textbf{RPR} & \textbf{RTC} & \textbf{SEI} & \textbf{Average} \\
\midrule
\multirow{5}{*}{DeepSeek V3} 
& DeepSeek R1            & 9.51 & \textbf{8.75} & \textbf{9.44} & \textbf{9.45} & \textbf{7.61} & \textbf{8.53} & \textbf{9.47} & 7.76 & \textbf{9.64} & \textbf{8.85} & \textbf{9.14} & \textbf{9.06} & \textbf{8.93} \\
& DeepSeek V3            & \textbf{9.57} & 8.61 & 9.25 & 9.27 & 7.23 & 7.98 & 9.21 & 7.56 & 8.94 & 8.76 & 9.00 & 8.59 & 8.66 \\
& Qwen Max               & 9.38 & 8.53 & 9.12 & 9.23 & 7.43 & 7.99 & 9.16 & \textbf{7.85} & 9.05 & 8.57 & 9.00 & 8.61 & 8.66 \\
& Qwen2.5-14B-Inst   & 9.28 & 8.50 & 9.03 & 9.14 & 7.14 & 7.81 & 8.94 & 7.55 & 8.71 & 8.35 & 8.82 & 8.25 & 8.46 \\
& Qwen2.5-7B-Inst    & 9.27 & 8.55 & 9.08 & 9.12 & 6.77 & 7.86 & 8.96 & 7.05 & 8.95 & 8.42 & 8.82 & 8.53 & 8.44 \\
\midrule
\multirow{5}{*}{Human}
& DeepSeek R1            & \textbf{8.97} & \textbf{8.60} & \textbf{8.98} & \textbf{8.94} & \textbf{8.86} & \textbf{8.56} & \textbf{8.77} & \textbf{8.20} & \textbf{9.26} & \textbf{7.95} & \textbf{8.91} & \textbf{8.92} & \textbf{8.74} \\
& DeepSeek V3            & 8.77 & 7.77 & 8.40 & 7.89 & 8.11 & 7.25 & 8.10 & 7.70 & 7.42 & 7.03 & 7.80 & 7.47 & 7.89 \\
& Qwen Max               & 8.81 & 8.01 & 8.52 & 8.27 & 8.23 & 7.59 & 8.10 & 7.70 & 7.89 & 7.31 & 8.09 & 7.74 & 8.02 \\
& Qwen2.5-14B-Inst   & 8.74 & 7.76 & 8.26 & 7.79 & 7.86 & 6.88 & 7.77 & 6.97 & 7.02 & 7.01 & 7.59 & 7.03 & 7.56 \\
& Qwen2.5-7B-Inst    & 8.49 & 7.63 & 8.04 & 7.82 & 7.45 & 6.93 & 7.65 & 7.05 & 7.38 & 5.90 & 7.82 & 7.35 & 7.46 \\
\bottomrule
\end{tabular}
\end{adjustbox}
\caption{Metric-level evaluation results under model-based and human evaluators}
\end{subtable}
\hfill
\begin{subtable}[t]{0.31\linewidth}
\centering
\begin{adjustbox}{width=\linewidth}
\begin{tabular}{cl}
\toprule
\textbf{Abbreviation} & \textbf{Full Name} \\
\midrule
IFTC  & Instruction Following \& Task Completion \\
RTC   & Role \& Tone Consistency \\
CRSC  & Content Relevance \& Scope Control \\
SEI   & Scenario Element Integration \\
\midrule
BFA   & Basic Factual Accuracy \\
DKA   & Domain Knowledge Accuracy \\
RPR   & Reasoning Process Rigor \\
EICP  & Error Identification \& Correction Precision \\
\midrule
CSI   & Clarity, Simplicity \& Inspiration \\
MGP   & Motivation, Guidance \& Positive Feedback \\
PAS   & Personalization, Adaptation \& Learning Support \\
HOTS  & Higher-Order Thinking \& Skill Development \\
\bottomrule
\end{tabular}
\end{adjustbox}
\caption{Full names of evaluation metrics}
\label{tab:metric-abbreviations}
\end{subtable}
\caption{Metric-level average scores evaluated by DeepSeek V3 and human evaluators across different evaluation metrics. For brevity, metric names are abbreviated, with full definitions provided in the right subtable.}
\label{tab:metric-results-v3-human}
\end{table*}

\begin{table*}[t]
\centering
\begin{adjustbox}{width=0.93\textwidth}
\begin{tabular}{l|l|ccccccccc|c}
\toprule
\textbf{Evaluator} & \textbf{Model} & \textbf{Q\&A} & \textbf{PLS} & \textbf{EC} & \textbf{IP} & \textbf{AG} & \textbf{TMG} & \textbf{ES} & \textbf{QG} & \textbf{PCC} & \textbf{Average} \\
\midrule
\multirow{5}{*}{DeepSeek V3}
& DeepSeek R1            & 9.49 & \textbf{9.65} & \textbf{9.27} & \textbf{8.75} & \textbf{7.27} & \textbf{9.45} & \textbf{9.38} & \textbf{9.33} & \textbf{9.71} & \textbf{9.14} \\
& DeepSeek V3            & \textbf{9.68} & 9.04 & 9.14 & 8.53 & 7.05 & 9.34 & 9.00 & 9.06 & 8.92 & 8.86 \\
& Qwen Max               & 9.18 & 8.88 & 9.06 & 8.52 & 7.23 & 9.24 & 9.04 & 9.05 & 9.29 & 8.83 \\
& Qwen2.5-14B-Instruct   & 9.07 & 8.72 & 8.97 & 8.30 & 6.77 & 9.21 & 8.74 & 9.02 & 8.80 & 8.62 \\
& Qwen2.5-7B-Instruct    & 9.15 & 9.07 & 9.01 & 8.47 & 6.44 & 9.21 & 8.85 & 8.69 & 9.00 & 8.65 \\
\midrule
\multirow{5}{*}{Human}
& DeepSeek R1            & 7.17 & \textbf{9.11} & \textbf{8.71} & \textbf{8.80} & \textbf{8.42} & \textbf{8.86} & \textbf{9.15} & \textbf{8.79} & \textbf{9.35} & \textbf{8.71} \\
& DeepSeek V3            & 7.45 & 8.12 & 8.16 & 8.17 & 7.84 & 7.56 & 8.08 & 8.01 & 7.03 & 7.82 \\
& Qwen Max               & \textbf{7.72} & 7.94 & 8.21 & 8.15 & 7.89 & 7.99 & 7.85 & 8.39 & 8.42 & 8.06 \\
& Qwen2.5-14B-Instruct   & 7.66 & 7.38 & 7.92 & 7.56 & 7.55 & 7.84 & 7.31 & 7.91 & 7.36 & 7.61 \\
& Qwen2.5-7B-Instruct    & 6.78 & 7.63 & 7.93 & 7.74 & 6.79 & 7.86 & 7.79 & 7.55 & 7.42 & 7.50 \\
\bottomrule
\end{tabular}
\end{adjustbox}
\caption{Scenario-Level average scores evaluated by DeepSeek V3 and human evaluator. Max values in each column per evaluator are bolded. Full names of each scenarios can be found in Section~\ref{sec:domain_design}.}
\label{tab:task-results-v3-human}
\end{table*}

\section{Experiments}

\paragraph{Response generation} 
We select 5 representative models: DeepSeek R1~\citep{guo2025deepseek}, DeepSeek V3~\citep{liu2024deepseek}, Qwen Max, Qwen2.5-14B-Instruct, and Qwen2.5-7B-Instruct~\citep{qwen25}. This selection provides a broad view of how models of varying sizes and types, such as standard and reasoning models, handle educational tasks.


\paragraph{Response evaluation} We select the following LLMs: QwQ Plus~\citep{qwq32b}, GPT-4o~\citep{openai2024gpt04o}, DeepSeek R1~\citep{guo2025deepseek}, and DeepSeek V3~\citep{liu2024deepseek}, as evaluators due to their strong scenario understanding, broad knowledge, and accurate intent recognition. These models assess responses using our defined metrics, guided by dedicated prompts that include language-specific descriptions to avoid multilingual bias. Prompt details are provided in Appendix~\ref{sec:eval-prompt}.


\paragraph{Test set} We use 198 test samples (99 Chinese, 99 English), comprising questions from Section \ref{sec:question-gen} and responses from five models. For data selection of the test set, we sample all the data points from different educational sub-contexts to ensure the diversity and comprehensiveness of our evaluation.

We present the evaluation results from the best-performing evaluator model: Deepseek V3, while the complete results of model generation and evaluation are detailed in Appendix~\ref{sec:extra-results}.
%

\subsection{Evaluation Details} 
As mentioned in Section~\ref{sec:human_guided}, to ensure the rationality and verifiability of the evaluation, we employ both model-based and human-based point-wise evaluations to assess the responses from different models.  
Specifically, each QA pair is evaluated separately by each evaluation model and by 3 human annotators. All evaluations are based on the 12 metrics that we have defined in Section~\ref{sec:metric-design}. 

During model evaluation, metric information is embedded into the prompt, requiring models to output individual metric scores in a single response. In human evaluation, annotators study the metrics in advance and adhere strictly to the criteria during annotation. We adopt a pointwise evaluation strategy, as preliminary experiments reveal significant positional bias in pairwise settings (Appendix~\ref{sec:pos_pref}). The scoring guidelines are detailed in Appendix~\ref{app:metric-scores}.


\subsection{Experiment Results}
\label{sec:overall_results}



\paragraph{Model-evaluation results} 
DeepSeek R1 demonstrates the best overall performance across different metrics, while Qwen2.5-7B-Instruct performs the worst in Table~\ref{tab:metric-results-v3-human}. Moreover, DeepSeek R1 performs the best on Higher-Order Thinking \& Skill Development, and Qwen2.5-7B-Instruct is the least satisfactory in Error Identification \& Correction Precision, with both models showing a clear gap compared to others. In specific scenarios, DeepSeek R1 remains the best, while Qwen2.5-7B-Instruct outperforms Qwen2.5-14B-Instruct in scenarios like Emotional Support and Personalized Content Creation in Table~\ref{tab:task-results-v3-human}. 
This shows the gap between models with smaller sizes is not very large and it drives us to choose the 7B model as the student model during our distillation experiments in Section~\ref{sec:distillation}.


\paragraph{Human-evaluation results} DeepSeek R1 and Qwen2.5-7B-Instruct still demonstrate the best and worst performance respectively in Table~\ref{tab:metric-results-v3-human}, consistent with the results from model-based evaluation. Unlike model evaluation, human annotation shows noticeably lower satisfaction with the performance of all five models on the Reasoning Process Rigor metric. Qwen2.5-7B-Instruct performs particularly poorly on this metric, scoring only 5.90. In contrast, DeepSeek R1 shows consistently strong performance on the Motivation, Guidance \& Positive Feedback metric, even when other models fall short. At the scenario level, DeepSeek R1 remains far ahead in Table~\ref{tab:task-results-v3-human}, while the performance gap between 7B and 14B of Qwen models is relatively small, making the 7B model a cost-effective choice in resource-constrained settings.


\begin{table*}[h]
\centering
\begin{adjustbox}{width=0.93\textwidth}
\begin{tabular}{l|cccccccccccc|c}
\toprule
\textbf{Model} & \textbf{BFA} & \textbf{CSI} & \textbf{CRSC} & \textbf{DKA} & \textbf{EICP} & \textbf{HOTS} & \textbf{IFTC} & \textbf{MGP} & \textbf{PAS} & \textbf{RPR} & \textbf{RTC} & \textbf{SEI} & \textbf{Average} \\
\midrule
DeepSeek R1            & \underline{9.51} & \textbf{8.75} & \textbf{9.44} & \textbf{9.45} & \textbf{7.61} & \textbf{8.53} & \textbf{9.47} & \underline{7.76} & \textbf{9.64} & \underline{8.85} & \textbf{9.14} & \textbf{9.06} & \textbf{8.93} \\
DeepSeek V3            & \textbf{9.57} & \underline{8.61} & 9.25 & \underline{9.27} & 7.23 & 7.98 & 9.21 & 7.56 & 8.94 & 8.76 & 9.00 & 8.59 & 8.66 \\
Qwen Max               & 9.38 & 8.53 & 9.12 & 9.23 & \underline{7.43} & 7.99 & 9.16 & \textbf{7.85} & 9.05 & 8.57 & 9.00 & 8.61 & 8.66 \\
Qwen2.5-14B-Instruct   & 9.28 & 8.50 & 9.03 & 9.14 & 7.14 & 7.81 & 8.94 & 7.55 & 8.71 & 8.35 & 8.82 & 8.25 & 8.46 \\
Qwen2.5-7B-Instruct    & 9.27 & 8.55 & 9.08 & 9.12 & 6.77 & 7.86 & 8.96 & 7.05 & 8.95 & 8.42 & 8.82 & 8.53 & 8.44 \\
Distillation Qwen2.5-7B       & 9.26 & 8.56 & \underline{9.27} & 8.95 & 6.89 & \underline{8.43} & \underline{9.41} & 7.32 & \underline{9.56} & \textbf{9.26} & \underline{9.09} & \underline{8.95} & \underline{8.75} \\
\bottomrule
\end{tabular}
\end{adjustbox}
\caption{Performance comparison of our distillation model with other models based on metric-level evaluations by DeepSeek V3. Best results are bold while second best results are underlined. For our distillation model, we use Qwen2.5-7B-Instruct as the base model. For simplicity, metrics are abbreviated; full names are listed in Table~\ref{tab:metric-abbreviations}.}
\label{tab:metric-results_distilled}
\end{table*}

\begin{table}[h]
\centering
\begin{adjustbox}{width=\linewidth}
\begin{tabular}{c|cccc|c}
\toprule
\textbf{Model}&\textbf{DeepSeek R1} & \textbf{GPT-4o} & \textbf{QwQ-Plus} & \textbf{DeepSeek V3} & \textbf{Human}\\
\midrule
\textbf{DeepSeek R1} &-&0.55&0.61&\textbf{0.65}&0.63\\
 \textbf{GPT-4o} &0.55&-&0.57&\textbf{0.58}&0.56\\
 \textbf{QwQ-Plus} &0.61&0.57&-&\textbf{0.62}&0.63\\
 \textbf{DeepSeek V3} &0.65&0.58&0.62&-&0.63\\
 \midrule
 \textbf{Human} &0.63&0.56&0.63&\textbf{0.63}&-\\
\bottomrule
\end{tabular}
\end{adjustbox}
\caption{\textbf{Kendall’s} $W$ between different evaluation models and human evaluation.}
\label{tab:consistency}
\end{table}

\subsection{Analysis}
\subsubsection{Consistency Results}
\paragraph{Consistency between evaluation models}  As shown in Table~\ref{tab:consistency}, Kendall’s W values exceed 0.5 for nearly all models, indicating strong agreement. DeepSeek V3 exhibits the highest consistency, with its average-score rankings closely aligned with those of other models. 

\noindent\textbf{Consistency between human and models} 
The overall rankings of the generation models from both human evaluation and model evaluation show similar trends. We also evaluated Kendall’s Coefficient of Concordance between different evaluators in Table~\ref{tab:consistency}.
The evaluation scores from models do not align precisely with human judgments, which may be attributed to their limited evaluation metric understanding. Overall, DeepSeek V3 exhibits the highest correlation with human evaluations, while GPT-4o shows the lowest. This pattern may be attributed to the relatively larger model sizes and broad training data distribution of DeepSeek V3.

\subsubsection{Model Behavior Analysis}
\paragraph{Model evaluations tend to assign higher scores than human annotators.} Our results demonstrate that models assign scores approximately one point higher than human's at both metric and scenario levels. Models typically assign scores exceeding 9, while human annotators generally label the same items with scores around 7, creating an almost two-point discrepancy that highlights a significant divergence in evaluation standards. We attribute this gap to two factors: 1) Models may misinterpret scoring criteria, which could be improved through post-training, as current evaluators are not reward models. 2) Reinforcement Learning from Human Feedback training makes models reluctant to give negative feedback. However, we believe post-training could mitigate this issue, and we'll explore this direction in future work.


\noindent\textbf{Larger models typically outperform smaller ones across scenarios.} Top models like DeepSeek R1 excel overall, while smaller ones (e.g., Qwen2.5-7B-Instruct) only succeed in limited tasks and lag in complex metrics (Domain Knowledge Accuracy, etc.). Notably, for smaller models, size is not definitive – the 7B Qwen2.5 occasionally surpasses its 14B version.


\section{Multi-source Distillation}
\label{sec:distillation}

\noindent\textbf{Data selection based on EduBench.}
To fully leverage the strengths of different response generation models across various scenarios, we adopt a multi-source distillation pipeline. For each task, we select the best-performing model on the test set as the response generator, using it to answer educational domain questions and construct the training dataset for the distillation model, details are shown in Appendix~\ref{app:data_selection}. Through the distillation pipeline, we obtain a training set of 17,000 samples covering various subtasks across all 9 educational scenarios.

\noindent\textbf{Results.}
As shown in Table~\ref{tab:metric-results_distilled}, after distillation, the performance of the 7B model significantly improved across 10 of the 12 metrics, achieving performance comparable to that of state-of-the-art models. Notably, it outperforms all other models, including DeepSeek R1 and Qwen Max, in terms of the Reasoning Process Rigor metric.

\section{Related Work}
\paragraph{Evaluation Benchmarks of LLMs}


Mainstream evaluation benchmarks for LLMs often prioritize the assessment of general capabilities rather than their applicability in educational contexts.
This gap has spurred specialized benchmarks that target educational scenarios, although significant limitations persist.
For instance, EduNLP~\citep{huang_edunlp_2024} offers a modular framework spanning 8 disciplines and 5 downstream tasks, but its pedagogical relevance is limited by narrow scenario diversity and reliance on generic metrics. 
In programming education, benchmarks on code repair evaluate LLMs in tasks such as code correction and error explanation across classroom and open-source settings~\cite{koutcheme_using_2024,koutcheme_benchmarking_2024}. In contrast, benchmarks for academic question answering and language learning focus on reasoning and text simplification across many domains, but typically adopt fixed formats that emphasize readability or correctness over interactive pedagogy~\cite{rein_gpqa_2023,team_supergpqa_2025,huangGeneratingEducationalMaterials2024a}.

Despite progress in general and domain-specific benchmarks, educational applications remain constrained to narrow scenarios, typically focusing on single-subject, fact-based Q\&A, while overlooking authentic contexts such as adaptive feedback and collaborative learning.
\paragraph{LLMs for Education Applications}
A substantial body of work investigates LLMs for educational applications across multiple settings, including content generation, interactive tutoring, and automated assessment. Prior studies show that LLMs enable scalable production of coherent instructional materials~\citep{jiaEQGRACEExaminationTypeQuestion2021,ghanemQuestionGenerationReading2022a,bermanLLMGeneratedMultipleChoice2024}, support learning in domains such as programming and mathematics through guided instruction and personalized feedback, and support instructional practice by simulating student behaviors or providing instructional feedback~\citep{macneilExperiencesUsingCode2023,wangEnhancingComputerProgramming2024a,pardosLearningGainDifferences2023,markelGPTeachInteractiveTA2023}. LLMs have also been widely adopted as automated assessment tools for evaluating essays and mathematical answers, offering scoring and diagnostic feedback to support learning~\citep{caoDomainAdaptiveNeuralAutomated2020,yangEnhancingAutomatedEssay2020,jiangLLMsCanFind2024}, with many evaluation settings further relying on option-based formats~\cite{rein_gpqa_2023}. In terms of evaluation and task design, existing studies commonly adopt surface-level metrics such as ROUGE or accuracy~\citep{DBLP:conf/acl/ShiLX25} and typically focus on classroom-only or Examination-oriented scenarios, rather than modeling broader and more diverse educational processes~\citep{DBLP:journals/corr/abs-2510-04648}.

Compared with prior work that focuses on Examination-oriented settings and rigid multiple-choice evaluations, our approach emphasizes both scenario-specific and diverse evaluation criteria, while also expanding the range of educational scenarios. As a result, our work covers a broad spectrum of educational contexts and supports multiple user groups within the educational domain.

\section{Discussion}


We believe that our work lays a foundational step for LLM-based educational research, offering a comprehensive benchmark and evaluation framework that captures the \textbf{diverse roles, scenarios, and needs} present in real-world education. By systematically incorporating scenarios like psychological counseling, assignment grading, and personalized study planning, we bring previously overlooked scenarios into the research landscape, encouraging deeper exploration across subject areas, learner profiles, and task types. Our benchmark can serve as a springboard for future research in designing better \textbf{better benchmarks that fulfill more diverse needs}, \textbf{robust evaluation models}, \textbf{scenario-adapted educational LLMs}, and even \textbf{LLM agents} that can perform multi-role, interactive support in classrooms or digital learning environments. 
Moreover, this work has immediate \textbf{practical value for educators and institutions}, offering structured tools that can help enhance efficiency, personalize learning, and reduce workload. The synthetic data construction methods we employ also open up new possibilities for \textbf{scalable, low-cost training and evaluation}, though future work could further improve context richness, realism, and dynamic data generation. Ultimately, we hope this work inspires the community to build stronger foundation models for trustworthy and effective educational AI systems.

\section{Conclusion}
In this work, we present the first comprehensive benchmark for evaluating LLMs in diverse educational scenarios. By incorporating data across 9 major domains and over 4,000 distinct educational contexts, it integrates model-generated query data to reflect real-world needs.  We further introduce a set of multi-dimensional evaluation metrics spanning 12 critical aspects, addressing the perspectives of both educators and learners. Human annotations are employed to validate the quality and relevance of model-generated outputs, enhancing the benchmark’s reliability. Extensive experiments show that smaller models trained on our dataset can rival state-of-the-art LLMs, underscoring the potential for efficient education-oriented LLMs. 
We believe this benchmark could serve as a valuable resource for the community and inspire further research in optimizing LLMs for educational applications.

\section*{Limitations}
This work has several limitations that point to promising directions for future research. First, the set of LLMs we evaluated is relatively limited, and including a wider variety of models would offer a more comprehensive understanding of system performance. 
Second, while our work explores the correlation between human and model evaluations, there is still room to improve alignment. We also employed only basic prompt engineering techniques; more sophisticated prompting strategies or the use of LLM agents may lead to better results. Moreover, most of our evaluation metrics and task scenarios were manually designed, and automating this process could enhance scalability and consistency. Finally, our methods have not yet been tested in real-world educational environments with practitioners, which will be important for validating practical applicability and impact.

\section*{Author Contributions}

\begin{itemize}
    \item \textbf{Bin Xu}: Conceived and designed the analysis; collected the data; performed the analysis; wrote the paper.
    \item \textbf{Yu Bai}: Developed the idea; conceived and designed the analysis; manage the project; performed the analysis; wrote the paper.
    \item \textbf{Huashan Sun}: Conceived and designed the analysis; collected the data; performed the analysis; wrote the paper.
    \item \textbf{Yiguan Lin}: Conceived and designed the analysis; collected the data; performed the analysis; wrote the paper.
    \item \textbf{Siming Liu}: Conceived and designed the analysis; performed the analysis; wrote the paper; provided training to the annotation company.
    \item \textbf{Xinyue Liang}: Performed the analysis; wrote the paper.
    \item \textbf{Yaolin Li}: Wrote the paper.
    \item \textbf{Zhuangzhi Dong}: Wrote the paper; collected the data.
    \item \textbf{Jingren Zhang}: Collected the data.
    \item \textbf{Yufan Deng}: Collected the data.
    \item \textbf{Xinyu Zou}: Collected the data. 

    \item \textbf{Yang Gao}: Conceptualization; lead and manage the project; provided computing resources and funding.
    \item \textbf{Heyan Huang}: Provided computing resources.
\end{itemize}

\section*{Ethical considerations}
In this study, all experiments involving human annotators comply fully with ethical standards set by a professional annotation company. Annotators participate voluntarily and are fully informed about the experimental procedures and task requirements before starting. Clear guidelines are provided, and annotators receive sufficient training to ensure consistency and fairness in the annotation process. All data remain anonymized, and the privacy of annotators is strictly protected. This study ensures that no tasks involve potential harm or ethical concerns, and all experiments follow relevant ethical guidelines.
\bibliography{custom}

\begin{thebibliography}{35}
\providecommand{\natexlab}[1]{#1}

\bibitem[{Bai et~al.(2022)Bai, Kadavath, Kundu, Askell, Kernion, Jones, Chen, Goldie, Mirhoseini, McKinnon et~al.}]{bai2022constitutional}
Yuntao Bai, Saurav Kadavath, Sandipan Kundu, Amanda Askell, Jackson Kernion, Andy Jones, Anna Chen, Anna Goldie, Azalia Mirhoseini, Cameron McKinnon, and 1 others. 2022.
\newblock Constitutional ai: Harmlessness from ai feedback.
\newblock \emph{arXiv preprint arXiv:2212.08073}.

\bibitem[{Berman et~al.(2024)Berman, McCoy, and Camarata}]{bermanLLMGeneratedMultipleChoice2024}
Jonathan Berman, Lise McCoy, and Troy Camarata. 2024.
\newblock \href {https://doi.org/10.1152/physiol.2024.39.S1.376} {{{LLM-Generated Multiple Choice Practice Quizzes}} for {{Pre-Clinical Medical Students}}; {{Use}} and {{Validity}}}.
\newblock \emph{Physiology}, 39(S1):376.

\bibitem[{Cao et~al.(2020)Cao, Jin, Wan, and Yu}]{caoDomainAdaptiveNeuralAutomated2020}
Yue Cao, Hanqi Jin, Xiaojun Wan, and Zhiwei Yu. 2020.
\newblock \href {https://doi.org/10.1145/3397271.3401037} {Domain-{{Adaptive Neural Automated Essay Scoring}}}.
\newblock In \emph{Proceedings of the 43rd {{International ACM SIGIR Conference}} on {{Research}} and {{Development}} in {{Information Retrieval}}}, pages 1011--1020, Virtual Event China. ACM.

\bibitem[{Cobbe et~al.(2021)Cobbe, Kosaraju, Bavarian, Chen, Jun, Kaiser, Plappert, Tworek, Hilton, Nakano, Hesse, and Schulman}]{cobbe2021trainingverifierssolvemath}
Karl Cobbe, Vineet Kosaraju, Mohammad Bavarian, Mark Chen, Heewoo Jun, Lukasz Kaiser, Matthias Plappert, Jerry Tworek, Jacob Hilton, Reiichiro Nakano, Christopher Hesse, and John Schulman. 2021.
\newblock \href {https://arxiv.org/abs/2110.14168} {Training verifiers to solve math word problems}.
\newblock \emph{Preprint}, arXiv:2110.14168.

\bibitem[{Gan et~al.(2023)Gan, Qi, Wu, and Lin}]{gan_large_2023}
Wensheng Gan, Zhenlian Qi, Jiayang Wu, and Jerry Chun-Wei Lin. 2023.
\newblock \href {https://doi.org/10.48550/arXiv.2311.13160} {Large {Language} {Models} in {Education}: {Vision} and {Opportunities}}.
\newblock \emph{arXiv preprint}.
\newblock Issue: arXiv:2311.13160 arXiv:2311.13160 [cs].

\bibitem[{Ghanem et~al.(2022)Ghanem, Lutz~Coleman, Rivard~Dexter, Von Der~Ohe, and Fyshe}]{ghanemQuestionGenerationReading2022a}
Bilal Ghanem, Lauren Lutz~Coleman, Julia Rivard~Dexter, Spencer Von Der~Ohe, and Alona Fyshe. 2022.
\newblock \href {https://doi.org/10.18653/v1/2022.findings-acl.168} {Question {{Generation}} for {{Reading Comprehension Assessment}} by {{Modeling How}} and {{What}} to {{Ask}}}.
\newblock In \emph{Findings of the {{Association}} for {{Computational Linguistics}}: {{ACL}} 2022}, pages 2131--2146, Dublin, Ireland. Association for Computational Linguistics.

\bibitem[{Guo et~al.(2025)Guo, Yang, Zhang, Song, Zhang, Xu, Zhu, Ma, Wang, Bi et~al.}]{guo2025deepseek}
Daya Guo, Dejian Yang, Haowei Zhang, Junxiao Song, Ruoyu Zhang, Runxin Xu, Qihao Zhu, Shirong Ma, Peiyi Wang, Xiao Bi, and 1 others. 2025.
\newblock Deepseek-r1: Incentivizing reasoning capability in llms via reinforcement learning.
\newblock \emph{arXiv preprint arXiv:2501.12948}.

\bibitem[{Huang et~al.(2024{\natexlab{a}})Huang, Wei, and Huang}]{huangGeneratingEducationalMaterials2024a}
Chieh-Yang Huang, Jing Wei, and Ting-Hao~Kenneth Huang. 2024{\natexlab{a}}.
\newblock \href {https://doi.org/10.1145/3690712.3690718} {Generating {{Educational Materials}} with {{Different Levels}} of {{Readability}} using {{LLMs}}}.
\newblock In \emph{Proceedings of the {{Third Workshop}} on {{Intelligent}} and {{Interactive Writing Assistants}}}, pages 16--22, Honolulu HI USA. ACM.

\bibitem[{Huang et~al.(2024{\natexlab{b}})Huang, Ning, Qin, Tong, Xue, Xiao, Lin, Liu, Liu, Chen, and Wang}]{huang_edunlp_2024}
Zhenya Huang, Yuting Ning, Longhu Qin, Shiwei Tong, Shangzi Xue, Tong Xiao, Xin Lin, Jiayu Liu, Qi~Liu, Enhong Chen, and Shijing Wang. 2024{\natexlab{b}}.
\newblock \href {https://doi.org/10.48550/arXiv.2406.01276} {{EduNLP}: {Towards} a {Unified} and {Modularized} {Library} for {Educational} {Resources}}.
\newblock \emph{arXiv preprint}.
\newblock Issue: arXiv:2406.01276 arXiv:2406.01276 [cs].

\bibitem[{Jia et~al.(2021)Jia, Zhou, Sun, and Wu}]{jiaEQGRACEExaminationTypeQuestion2021}
Xin Jia, Wenjie Zhou, Xu~Sun, and Yunfang Wu. 2021.
\newblock \href {https://doi.org/10.1609/aaai.v35i14.17553} {{{EQG-RACE}}: {{Examination-Type Question Generation}}}.
\newblock \emph{Proceedings of the AAAI Conference on Artificial Intelligence}, 35(14):13143--13151.

\bibitem[{Jiang et~al.(2024)Jiang, Peng, Feng, Li, and Li}]{jiangLLMsCanFind2024}
Zhuoxuan Jiang, Haoyuan Peng, Shanshan Feng, Fan Li, and Dongsheng Li. 2024.
\newblock \href {https://doi.org/10.48550/ARXIV.2405.06705} {{{LLMs}} can {{Find Mathematical Reasoning Mistakes}} by {{Pedagogical Chain-of-Thought}}}.
\newblock \emph{arXiv preprint}.

\bibitem[{Koutcheme et~al.(2024{\natexlab{a}})Koutcheme, Dainese, and Hellas}]{koutcheme_using_2024}
Charles Koutcheme, Nicola Dainese, and Arto Hellas. 2024{\natexlab{a}}.
\newblock \href {https://aclanthology.org/2024.bea-1.15/} {Using {Program} {Repair} as a {Proxy} for {Language} {Models}' {Feedback} {Ability} in {Programming} {Education}}.
\newblock In \emph{Proceedings of the 19th {Workshop} on {Innovative} {Use} of {NLP} for {Building} {Educational} {Applications} ({BEA} 2024)}, pages 165--181, Mexico City, Mexico. Association for Computational Linguistics.

\bibitem[{Koutcheme et~al.(2024{\natexlab{b}})Koutcheme, Dainese, Sarsa, Leinonen, Hellas, and Denny}]{koutcheme_benchmarking_2024}
Charles Koutcheme, Nicola Dainese, Sami Sarsa, Juho Leinonen, Arto Hellas, and Paul Denny. 2024{\natexlab{b}}.
\newblock \href {https://doi.org/10.48550/arXiv.2405.05347} {Benchmarking {Educational} {Program} {Repair}}.
\newblock \emph{arXiv preprint}.
\newblock Issue: arXiv:2405.05347 arXiv:2405.05347 [cs].

\bibitem[{Li et~al.(2025)Li, Gao, Yang, Bai, Zhou, Li, Sun, Liu, Si, Ye et~al.}]{li2025fundamental}
Jiawei Li, Yang Gao, Yizhe Yang, Yu~Bai, Xiaofeng Zhou, Yinghao Li, Huashan Sun, Yuhang Liu, Xingpeng Si, Yuhao Ye, and 1 others. 2025.
\newblock Fundamental capabilities and applications of large language models: A survey.
\newblock \emph{ACM Computing Surveys}.

\bibitem[{Liu et~al.(2024)Liu, Feng, Xue, Wang, Wu, Lu, Zhao, Deng, Zhang, Ruan et~al.}]{liu2024deepseek}
Aixin Liu, Bei Feng, Bing Xue, Bingxuan Wang, Bochao Wu, Chengda Lu, Chenggang Zhao, Chengqi Deng, Chenyu Zhang, Chong Ruan, and 1 others. 2024.
\newblock Deepseek-v3 technical report.
\newblock \emph{arXiv preprint arXiv:2412.19437}.

\bibitem[{Liu et~al.(2025)Liu, Wang, Xu, Ma, Ruan, Li, Liu, and Wu}]{liu2025inference}
Zijun Liu, Peiyi Wang, Runxin Xu, Shirong Ma, Chong Ruan, Peng Li, Yang Liu, and Yu~Wu. 2025.
\newblock Inference-time scaling for generalist reward modeling.
\newblock \emph{arXiv preprint arXiv:2504.02495}.

\bibitem[{MacNeil et~al.(2023)MacNeil, Tran, Hellas, Kim, Sarsa, Denny, Bernstein, and Leinonen}]{macneilExperiencesUsingCode2023}
Stephen MacNeil, Andrew Tran, Arto Hellas, Joanne Kim, Sami Sarsa, Paul Denny, Seth Bernstein, and Juho Leinonen. 2023.
\newblock \href {https://doi.org/10.1145/3545945.3569785} {Experiences from {{Using Code Explanations Generated}} by {{Large Language Models}} in a {{Web Software Development E-Book}}}.
\newblock In \emph{Proceedings of the 54th {{ACM Technical Symposium}} on {{Computer Science Education V}}. 1}, pages 931--937, Toronto ON Canada. ACM.

\bibitem[{Markel et~al.(2023)Markel, Opferman, Landay, and Piech}]{markelGPTeachInteractiveTA2023}
Julia~M. Markel, Steven~G. Opferman, James~A. Landay, and Chris Piech. 2023.
\newblock \href {https://doi.org/10.1145/3573051.3593393} {{{GPTeach}}: {{Interactive TA Training}} with {{GPT-based Students}}}.
\newblock In \emph{Proceedings of the {{Tenth ACM Conference}} on {{Learning}} @ {{Scale}}}, pages 226--236, Copenhagen Denmark. ACM.

\bibitem[{Ng and Fung(2024)}]{ng_educational_2024}
Chee Ng and Yuen Fung. 2024.
\newblock \href {https://doi.org/10.48550/arXiv.2407.11773} {Educational {Personalized} {Learning} {Path} {Planning} with {Large} {Language} {Models}}.
\newblock \emph{arXiv preprint}.
\newblock Issue: arXiv:2407.11773 arXiv:2407.11773 [cs].

\bibitem[{OpenAI et~al.(2024)OpenAI, :, Hurst, Lerer, Goucher, Perelman, Ramesh, Clark, Ostrow, Welihinda, Hayes, Radford, Mądry, Baker-Whitcomb, Beutel, Borzunov, Carney, Chow, Kirillov, Nichol, Paino, Renzin, Passos, Kirillov, Christakis, Conneau, Kamali, Jabri, Moyer, Tam, Crookes, Tootoochian, Tootoonchian, Kumar, Vallone, Karpathy, Braunstein, Cann, Codispoti, Galu, Kondrich, Tulloch, Mishchenko, Baek, Jiang, Pelisse, Woodford, Gosalia, Dhar, Pantuliano, Nayak, Oliver, Zoph, Ghorbani, Leimberger, Rossen, Sokolowsky, Wang, Zweig, Hoover, Samic, McGrew, Spero, Giertler, Cheng, Lightcap, Walkin, Quinn, Guarraci, Hsu, Kellogg, Eastman, Lugaresi, Wainwright, Bassin, Hudson, Chu, Nelson, Li, Shern, Conger, Barette, Voss, Ding, Lu, Zhang, Beaumont, Hallacy, Koch, Gibson, Kim, Choi, McLeavey, Hesse, Fischer, Winter, Czarnecki, Jarvis, Wei, Koumouzelis, Sherburn, Kappler, Levin, Levy, Carr, Farhi, Mely, Robinson, Sasaki, Jin, Valladares, Tsipras, Li, Nguyen, Findlay, Oiwoh, Wong, Asdar, Proehl, Yang, Antonow,
  Kramer, Peterson, Sigler, Wallace, Brevdo, Mays, Khorasani, Such, Raso, Zhang, von Lohmann, Sulit, Goh, Oden, Salmon, Starace, Brockman, Salman, Bao, Hu, Wong, Wang, Schmidt, Whitney, Jun, Kirchner, de~Oliveira~Pinto, Ren, Chang, Chung, Kivlichan, O'Connell, O'Connell, Osband, Silber, Sohl, Okuyucu, Lan, Kostrikov, Sutskever, Kanitscheider, Gulrajani, Coxon, Menick, Pachocki, Aung, Betker, Crooks, Lennon, Kiros, Leike, Park, Kwon, Phang, Teplitz, Wei, Wolfe, Chen, Harris, Varavva, Lee, Shieh, Lin, Yu, Weng, Tang, Yu, Jang, Candela, Beutler, Landers, Parish, Heidecke, Schulman, Lachman, McKay, Uesato, Ward, Kim, Huizinga, Sitkin, Kraaijeveld, Gross, Kaplan, Snyder, Achiam, Jiao, Lee, Zhuang, Harriman, Fricke, Hayashi, Singhal, Shi, Karthik, Wood, Rimbach, Hsu, Nguyen, Gu-Lemberg, Button, Liu, Howe, Muthukumar, Luther, Ahmad, Kai, Itow, Workman, Pathak, Chen, Jing, Guy, Fedus, Zhou, Mamitsuka, Weng, McCallum, Held, Ouyang, Feuvrier, Zhang, Kondraciuk, Kaiser, Hewitt, Metz, Doshi, Aflak, Simens, Boyd,
  Thompson, Dukhan, Chen, Gray, Hudnall, Zhang, Aljubeh, Litwin, Zeng, Johnson, Shetty, Gupta, Shah, Yatbaz, Yang, Zhong, Glaese, Chen, Janner, Lampe, Petrov, Wu, Wang, Fradin, Pokrass, Castro, de~Castro, Pavlov, Brundage, Wang, Khan, Murati, Bavarian, Lin, Yesildal, Soto, Gimelshein, Cone, Staudacher, Summers, LaFontaine, Chowdhury, Ryder, Stathas, Turley, Tezak, Felix, Kudige, Keskar, Deutsch, Bundick, Puckett, Nachum, Okelola, Boiko, Murk, Jaffe, Watkins, Godement, Campbell-Moore, Chao, McMillan, Belov, Su, Bak, Bakkum, Deng, Dolan, Hoeschele, Welinder, Tillet, Pronin, Tillet, Dhariwal, Yuan, Dias, Lim, Arora, Troll, Lin, Lopes, Puri, Miyara, Leike, Gaubert, Zamani, Wang, Donnelly, Honsby, Smith, Sahai, Ramchandani, Huet, Carmichael, Zellers, Chen, Chen, Nigmatullin, Cheu, Jain, Altman, Schoenholz, Toizer, Miserendino, Agarwal, Culver, Ethersmith, Gray, Grove, Metzger, Hermani, Jain, Zhao, Wu, Jomoto, Wu, Shuaiqi, Xia, Phene, Papay, Narayanan, Coffey, Lee, Hall, Balaji, Broda, Stramer, Xu, Gogineni,
  Christianson, Sanders, Patwardhan, Cunninghman, Degry, Dimson, Raoux, Shadwell, Zheng, Underwood, Markov, Sherbakov, Rubin, Stasi, Kaftan, Heywood, Peterson, Walters, Eloundou, Qi, Moeller, Monaco, Kuo, Fomenko, Chang, Zheng, Zhou, Manassra, Sheu, Zaremba, Patil, Qian, Kim, Cheng, Zhang, He, Zhang, Jin, Dai, and Malkov}]{openai2024gpt04o}
OpenAI, :, Aaron Hurst, Adam Lerer, Adam~P. Goucher, Adam Perelman, Aditya Ramesh, Aidan Clark, AJ~Ostrow, Akila Welihinda, Alan Hayes, Alec Radford, Aleksander Mądry, Alex Baker-Whitcomb, Alex Beutel, Alex Borzunov, Alex Carney, Alex Chow, Alex Kirillov, and 401 others. 2024.
\newblock Gpt-4o system card.
\newblock \emph{arXiv preprint arXiv: 2410.21276}.

\bibitem[{Pardos and Bhandari(2023)}]{pardosLearningGainDifferences2023}
Zachary~A. Pardos and Shreya Bhandari. 2023.
\newblock \href {https://doi.org/10.48550/ARXIV.2302.06871} {Learning gain differences between {{ChatGPT}} and human tutor generated algebra hints}.
\newblock \emph{arXiv preprint}.

\bibitem[{Qwen(2024)}]{qwen25}
Team Qwen. 2024.
\newblock Qwen2.5 technical report.
\newblock \emph{arXiv preprint arXiv:2412.15115}.

\bibitem[{Qwen(2025)}]{qwq32b}
Team Qwen. 2025.
\newblock \href {https://qwenlm.github.io/blog/qwq-32b/} {Qwq-32b: Embracing the power of reinforcement learning}.

\bibitem[{Rein et~al.(2023)Rein, Hou, Stickland, Petty, Pang, Dirani, Michael, and Bowman}]{rein_gpqa_2023}
David Rein, Betty~Li Hou, Asa~Cooper Stickland, Jackson Petty, Richard~Yuanzhe Pang, Julien Dirani, Julian Michael, and Samuel~R. Bowman. 2023.
\newblock \href {https://doi.org/10.48550/arXiv.2311.12022} {{GPQA}: {A} {Graduate}-{Level} {Google}-{Proof} {Q}\&{A} {Benchmark}}.
\newblock \emph{arXiv preprint}.
\newblock Issue: arXiv:2311.12022 arXiv:2311.12022 [cs].

\bibitem[{Rouzegar and Makrehchi(2024)}]{rouzegarGenerativeAIEnhancing2024}
Hamdireza Rouzegar and Masoud Makrehchi. 2024.
\newblock \href {https://doi.org/10.48550/ARXIV.2406.13903} {Generative {{AI}} for {{Enhancing Active Learning}} in {{Education}}: {{A Comparative Study}} of {{GPT-3}}.5 and {{GPT-4}} in {{Crafting Customized Test Questions}}}.

\bibitem[{Sharma et~al.(2025)Sharma, Tong, Mu, Wei, Kruthoff, Goodfriend, Ong, Peng, Agarwal, Anil et~al.}]{sharma2025constitutional}
Mrinank Sharma, Meg Tong, Jesse Mu, Jerry Wei, Jorrit Kruthoff, Scott Goodfriend, Euan Ong, Alwin Peng, Raj Agarwal, Cem Anil, and 1 others. 2025.
\newblock Constitutional classifiers: Defending against universal jailbreaks across thousands of hours of red teaming.
\newblock \emph{arXiv preprint arXiv:2501.18837}.

\bibitem[{Shi et~al.(2025)Shi, Liang, and Xu}]{DBLP:conf/acl/ShiLX25}
Yao Shi, Rongkeng Liang, and Yong Xu. 2025.
\newblock \href {https://aclanthology.org/2025.acl-long.1576/} {Educationq: Evaluating llms' teaching capabilities through multi-agent dialogue framework}.
\newblock In \emph{Proceedings of the 63rd Annual Meeting of the Association for Computational Linguistics (Volume 1: Long Papers), {ACL} 2025, Vienna, Austria, July 27 - August 1, 2025}, pages 32799--32828. Association for Computational Linguistics.

\bibitem[{Team et~al.(2025)Team, Du, Yao, Ma, Wang, Zheng, Zhu, Liu, Liang, Jin, Wei, Zheng, Deng, Gavin, Jia, Jiang, Liao, Li, Li, Li, Li, Li, Ma, Ni, Que, Wang, Wen, Wu, Hsing, Xu, Yang, Wang, Zhou, Bai, Bu, Cai, Chen, Chen, Cheng, Cheng, Ding, Huang, Huang, Li, Li, Li, Liang, Lin, Lin, Ma, Pang, Peng, Peng, Qi, Qiu, Qu, Quan, Tan, Wang, Wang, Wang, Wang, Wang, Xu, Yang, Yuan, Yue, Zhan, Zhang, Zhang, Zhang, Zhang, Zhang, Zhao, Zheng, Zhong, Gao, Li, Liu, Liu, Liu, Ni, Peng, Qin, Su, Wang, Wang, Yang, Yang, Cao, Yue, Zhang, Zhou, Liu, Lin, Huang, and Zhang}]{team_supergpqa_2025}
M.-A.-P. Team, Xinrun Du, Yifan Yao, Kaijing Ma, Bingli Wang, Tianyu Zheng, King Zhu, Minghao Liu, Yiming Liang, Xiaolong Jin, Zhenlin Wei, Chujie Zheng, Kaixin Deng, Shawn Gavin, Shian Jia, Sichao Jiang, Yiyan Liao, Rui Li, Qinrui Li, and 78 others. 2025.
\newblock \href {https://doi.org/10.48550/arXiv.2502.14739} {{SuperGPQA}: {Scaling} {LLM} {Evaluation} across 285 {Graduate} {Disciplines}}.
\newblock \emph{arXiv preprint}.
\newblock Issue: arXiv:2502.14739 arXiv:2502.14739 [cs].

\bibitem[{Wang et~al.(2024{\natexlab{a}})Wang, Milani, Chiu, Zhi, Eack, Labrum, Murphy, Jones, Hardy, Shen et~al.}]{wang2024patient}
Ruiyi Wang, Stephanie Milani, Jamie~C Chiu, Jiayin Zhi, Shaun~M Eack, Travis Labrum, Samuel~M Murphy, Nev Jones, Kate Hardy, Hong Shen, and 1 others. 2024{\natexlab{a}}.
\newblock Patient-$\{$$\backslash$Psi$\}$: Using large language models to simulate patients for training mental health professionals.
\newblock \emph{arXiv preprint arXiv:2405.19660}.

\bibitem[{Wang et~al.(2024{\natexlab{b}})Wang, Xu, Li, Zhang, Liang, Tang, Yu, and Wen}]{wang_large_2024}
Shen Wang, Tianlong Xu, Hang Li, Chaoli Zhang, Joleen Liang, Jiliang Tang, Philip~S. Yu, and Qingsong Wen. 2024{\natexlab{b}}.
\newblock \href {https://doi.org/10.48550/arXiv.2403.18105} {Large {Language} {Models} for {Education}: {A} {Survey} and {Outlook}}.
\newblock \emph{arXiv preprint}.
\newblock Issue: arXiv:2403.18105 arXiv:2403.18105 [cs] version: 1.

\bibitem[{Wang et~al.(2024{\natexlab{c}})Wang, Zhou, and Chen}]{wangEnhancingComputerProgramming2024a}
Tianyu Wang, Nianjun Zhou, and Zhixiong Chen. 2024{\natexlab{c}}.
\newblock \href {https://doi.org/10.48550/ARXIV.2407.05437} {Enhancing {{Computer Programming Education}} with {{LLMs}}: {{A Study}} on {{Effective Prompt Engineering}} for {{Python Code Generation}}}.
\newblock \emph{arXiv preprint}.

\bibitem[{Yang et~al.(2024)Yang, Ziems, Held, Shaikh, Bernstein, and Mitchell}]{yang_social_2024}
Diyi Yang, Caleb Ziems, William Held, Omar Shaikh, Michael~S. Bernstein, and John Mitchell. 2024.
\newblock \href {https://doi.org/10.48550/arXiv.2404.04204} {Social {Skill} {Training} with {Large} {Language} {Models}}.
\newblock \emph{arXiv preprint}.
\newblock Issue: arXiv:2404.04204 arXiv:2404.04204 [cs].

\bibitem[{Yang et~al.(2020)Yang, Cao, Wen, Wu, and He}]{yangEnhancingAutomatedEssay2020}
Ruosong Yang, Jiannong Cao, Zhiyuan Wen, Youzheng Wu, and Xiaodong He. 2020.
\newblock \href {https://doi.org/10.18653/v1/2020.findings-emnlp.141} {Enhancing {{Automated Essay Scoring Performance}} via {{Fine-tuning Pre-trained Language Models}} with {{Combination}} of {{Regression}} and {{Ranking}}}.
\newblock In \emph{Findings of the {{Association}} for {{Computational Linguistics}}: {{EMNLP}} 2020}, pages 1560--1569, Online. Association for Computational Linguistics.

\bibitem[{Zhang et~al.(2025)Zhang, Zhang-Li, Yu, Gong, Zhou, Hao, Jiang, Cao, Liu, Liu, Hou, and Li}]{zhang_simulating_2025}
Zheyuan Zhang, Daniel Zhang-Li, Jifan Yu, Linlu Gong, Jinchang Zhou, Zhanxin Hao, Jianxiao Jiang, Jie Cao, Huiqin Liu, Zhiyuan Liu, Lei Hou, and Juanzi Li. 2025.
\newblock \href {https://aclanthology.org/2025.naacl-long.520/} {Simulating {Classroom} {Education} with {LLM}-{Empowered} {Agents}}.
\newblock In \emph{Proceedings of the 2025 {Conference} of the {Nations} of the {Americas} {Chapter} of the {Association} for {Computational} {Linguistics}: {Human} {Language} {Technologies} ({Volume} 1: {Long} {Papers})}, pages 10364--10379, Albuquerque, New Mexico. Association for Computational Linguistics.

\bibitem[{Zhu et~al.(2025)Zhu, Hu, Ma, Zhang, and Cheong}]{DBLP:journals/corr/abs-2510-04648}
Buyuan Zhu, Shiyu Hu, Yiping Ma, Yuanming Zhang, and Kang~Hao Cheong. 2025.
\newblock \href {https://doi.org/10.48550/ARXIV.2510.04648} {Edupersona: Benchmarking subjective ability boundaries of virtual student agents}.
\newblock \emph{CoRR}, abs/2510.04648.

\end{thebibliography}

\appendix
\label{sec:appendix}
\appendix

\section{Detailed Scenario Descriptions}
\label{appendix:domain-design}
This appendix provides comprehensive definitions and explanations of all task scenarios included in EduBench. Each scenario captures a specific educational use case, featuring unique roles, goals, and evaluation considerations. The descriptions are organized into two categories—student-oriented and teacher-oriented—based on the primary user group served by the tasks.
\subsection{Student-Oriented Scenarios}
This section outlines scenarios in which AI systems directly assist students in their learning journey. The tasks are designed to reflect authentic student needs, including solving academic problems, receiving feedback on errors, obtaining personalized learning recommendations, and even receiving emotional support. These scenarios emphasize interaction quality, content accuracy, and adaptability to individual learning situations.
\begin{itemize}
    \item \textbf{Problem Solving:} The ability of an AI system to accurately solve questions posed by students across various subjects and difficulty levels.
    \item \textbf{Error Correction:} The capacity to identify and correct student errors in assignments, exams, or daily exercises. Errors can range from obvious mistakes to subtle issues such as variable misuse in code or logical flaws in mathematical reasoning. Evaluation focuses on the accuracy of error detection and the quality of correction.
    \item \textbf{Idea Provision:} This includes answering student queries about knowledge points, homework guidance, or exam preparation. It is subdivided into basic factual explanations, step-by-step solution analysis, and general academic advice. Responses are evaluated for accuracy, clarity, and informativeness.
    \item \textbf{Personalized Learning Support:} Based on student profiles (e.g., skill level, learning goals), the system recommends learning paths, exercises, or reading materials tailored to individual needs. Effectiveness is judged by the relevance, difficulty alignment, and usefulness of the recommendations.
    \item \textbf{Emotional Support:} This involves detecting a student's emotional state (e.g., anxiety before exams) from text and offering appropriate supportive feedback or suggestions. Scenarios include pre-exam stress, post-exam frustration, or social isolation. Evaluation metrics include emotion classification accuracy, specificity of emotional cues, and quality of suggestions.
\end{itemize}

\subsection{Teacher-Oriented Scenarios}
This section focuses on scenarios where AI systems are used to support educators in instructional design, assessment, and personalized teaching. These tasks capture the typical responsibilities of teachers—such as generating exam questions, grading, and preparing learning materials—and evaluate how effectively AI can augment or automate these functions to improve teaching efficiency and quality.
\begin{itemize}
    \item \textbf{Question Generation:} Generating questions based on specified topics, difficulty levels, and knowledge scopes. This includes both single-topic and multi-topic (comprehensive) question generation. Advanced requirements involve generating explanations and formatting full exams. Evaluation focuses on question quality, relevance, and structural coherence.
    \item \textbf{Automatic Grading:} Supporting grading of objective questions (e.g., multiple-choice, fill-in-the-blank) and subjective tasks (e.g., project reports) based on scoring rubrics. Feedback generation is also supported. Metrics include scoring accuracy, reasonableness, and feedback informativeness.
    \item \textbf{Teaching Material Generation:} Automatically generating educational content such as slides, teaching plans, and lecture notes. This includes content structuring and supplementing with relevant external materials like images or references.
    \item \textbf{Personalized Content Creation:} Generating differentiated content for students based on their learning levels or personal profiles. This includes both individualized assignments and tiered content design (e.g., differentiated learning objectives, teaching strategies, and assessments for varying student levels). Evaluation focuses on the internal validity of each item and cross-tier consistency.
\end{itemize}

\section{Educational Context Details}
\label{appendix:educational-context}


\begin{description}

\item[Subject Taxonomy:]
We follow a two-tier classification system to reflect academic breadth:

\begin{itemize}
\item \textbf{K–12 Subjects:}
\begin{itemize}
\item Chinese, Mathematics, English, Physics, Chemistry, Biology, History, Geography.
\end{itemize}

\item \textbf{Higher Education Subjects:}  
\begin{itemize}
    \item \textbf{Sciences:} Mathematics, Physics, Chemistry, Biology, Astronomy
    \item \textbf{Engineering:} Computer Science, Automation Control, Aerospace Science and Technology
    \item \textbf{Agriculture:} Aquaculture, Crop Science
    \item \textbf{Economics:} Applied Economics, Theoretical Economics
    \item \textbf{Education:} General Education, Physical Education
    \item \textbf{Management:} Business Administration, Public Administration
    \item \textbf{Medicine:} Basic Medicine, Clinical Medicine
    \item \textbf{Social Sciences and Humanities:} Sociology, Psychology, History, Law, Management
    \item \textbf{Literature and Arts:} Linguistics, Journalism, Theory of Music
    \item \textbf{Military Science}
\end{itemize}

\end{itemize}

\item[Task Difficulty Design:]
Tasks are divided into three levels:
\begin{itemize}
\item \textbf{Easy} – Basic knowledge and low cognitive load.
\item \textbf{Medium} – Intermediate tasks requiring moderate understanding.
\item \textbf{Hard} – Complex problems demanding critical reasoning and expertise.
\end{itemize}

\item[Language:]
\begin{itemize}
\item EduBench currently supports tasks in \textbf{Chinese and English}.
\end{itemize}

\item[Question Types:]
The question type dimension captures the format of interaction or evaluation expected from the model:
\begin{itemize}
    \item \textbf{Standard Scenarios (e.g., Problem Solving, Idea Provision, Grading):}
    \begin{itemize}
        \item Single Choice
        \item Multiple Choice
        \item Short Answer
    \end{itemize}
    \item \textbf{Emotional Support Scenarios:}
    \begin{itemize}
        \item Mental Healthy
        \item Mild Anxiety
        \item Moderate Anxiety
        \item Severe Anxiety
    \end{itemize}
    \item \textbf{Personalized Learning Support \& Personalized Content Creation:}
    \begin{itemize}
        \item No explicit question type; task outputs are tailored recommendations or generated content based on learner profiles.
    \end{itemize}
\end{itemize}

\end{description}

\section{Query Quality Evaluation}
\label{appendix:quality-eval}

\begin{table*}[htbp]
    \centering
    \caption{Comparison of Evaluation Scores Across Different Tasks and Annotation Sources}
    \label{tab:query_evaluation}
    \begin{adjustbox}{width=0.8\textwidth}
    \begin{tabular}{l c | c c c}
        \toprule
        \textbf{Task Name} & 
        \textbf{Human Labeling} & 
        \multicolumn{3}{c}{\textbf{Large Language Models}} \\
         & 
        & \textbf{R1} & \textbf{GPT-4o} & \textbf{QwQ} \\
        \midrule
        Linguistic Quality & 9.73 & 8.96 & 9.12 & 9.31 \\
        Content Quality & 9.66 & 9.36 & 8.91 & 9.45 \\
        Error Correction & 8.26 & 8.92 & 8.13 & 9.38 \\
        Problem Solving & 9.07 & 9.03 & 8.37 & 9.11 \\
        Automatic Grading & 8.31 & 7.75 & 7.96 & 8.50 \\
        Teaching Material Generation & 6.98 & 7.41 & 7.74 & 8.81 \\
        Question Generation & 8.81 & 7.84 & 8.29 & 9.15 \\
        Emotional Support & 8.07 & 8.28 & 6.85 & 8.77 \\
        Personalized Content Creation & 8.17 & 7.70 & 6.71 & 8.78 \\
        Personalized Learning Support & 8.31 & 7.09 & 6.51 & 9.00 \\
        \bottomrule
    \end{tabular}
    \end{adjustbox}
\end{table*}

\section{Query Evaluation Principles}
\label{query-evaluation-principles}
\subsection{Linguistic Quality}
\paragraph{Description} Is the language expression of the Query text clear, natural, fluent, and grammatically correct? Are there any spelling, punctuation, or word usage errors?

\begin{itemize}
\item \textbf{9-10:} Professional and refined language expression, completely free of grammatical, spelling, or punctuation errors, with natural and fluent phrasing.
\item \textbf{7-8:} Clear language expression, generally fluent, with possibly very minor spelling or punctuation oversights that do not affect understanding.
\item \textbf{5-6:} Basically understandable, but contains some obvious grammatical errors, awkward word choices, or sentence structure issues, causing some difficulty in comprehension.
\item \textbf{3-4:} Incoherent language expression, containing multiple severe grammatical or spelling errors, resulting in ambiguous meaning.
\item \textbf{1-2:} Completely incoherent, grammatically chaotic, filled with errors, making it almost impossible to understand the true intent.
\end{itemize}

\subsection{Content Quality}
\paragraph{Description} Is the content contained in the Query logically self-consistent and relevant to the educational context?

\begin{itemize}
\item \textbf{9-10:} Content is entirely logically self-consistent and highly relevant to educational objectives.
\item \textbf{7-8:} Content is basically logically reasonable and strongly relevant to the educational context. There might be minor details that are not rigorous enough.
\item \textbf{5-6:} Content is largely relevant but contains some unimportant factual errors or minor logical flaws.
\item \textbf{3-4:} Content contains obvious logical contradictions or has low relevance to the educational context.
\item \textbf{1-2:} Content contains severe logical contradictions or is completely irrelevant to education.
\end{itemize}

\subsection{Error Correction Task Quality}
\paragraph{Description} How good is this Query as an error correction task? The core lies in whether the provided text contains a clear, typical, and worth-correcting error, thus constituting a valuable evaluation instance.

\begin{itemize}
\item \textbf{9-10:} Perfect example. Contains a clear, typical (commonly made by students) error with a definite correct answer, in a natural context, making it highly valuable for evaluation.
\item \textbf{7-8:} Good example. Contains a clear error, but the error might be less typical or too simple; still has high evaluation value.
\item \textbf{5-6:} Usable example. An error exists in the text, but the error point might be unclear, ambiguous, or the text itself is unnaturally written.
\item \textbf{3-4:} Low-quality example. The provided text has no error, or the error is fabricated/meaningless, offering little evaluation value.
\item \textbf{1-2:} Invalid example. No text requiring correction is provided, or the content cannot constitute an error correction task
\end{itemize}

\subsection{Problem Solving Task Quality}
\paragraph{Description} How good is this Query as a Q\&A task? The core lies in whether the question itself is clear, effective, knowledge-rich, and has an answerable answer.

\begin{itemize}
\item \textbf{9-10:} High-quality question. The question is clearly and explicitly stated, knowledge-intensive, has a clear, verifiable answer, and matches the target students' cognitive level.
\item \textbf{7-8:} Good question. The question is clear and answerable but might be too broad, too simple, or slightly common-sense.
\item \textbf{5-6:} Average question. The question is somewhat vague or relies on unprovided context, but its intent can be roughly guessed and an attempt to answer can be made.
\item \textbf{3-4:} Low-quality question. The question is based on incorrect assumptions, logically flawed, or is a highly subjective 'pseudo-question' with no standard answer.
\item \textbf{1-2:} Invalid question. Not a question, or complete nonsense; incomprehensible and unanswerable.
\end{itemize}

\subsection{Automatic Grading Task Quality}
\paragraph{Description} How good is this Query as a judging task (evaluating a student's answer)? The core lies in whether it completely contains the two elements: question and answer, and whether the judgment of the answer's correctness possesses a certain level of challenge and discriminative power.

\begin{itemize}
\item \textbf{9-10:} High-quality judging task. The student's answer represents a typical error worth analyzing or a partially correct answer worth evaluating.
\item \textbf{7-8:} Good judging task. The correctness or incorrectness of the student's answer is too obvious.
\item \textbf{5-6:} Usable judging task. The question is overly simple, or the student's answer is of low quality (e.g., random guessing).
\item \textbf{3-4:} Low-quality judging task. Lacks core elements, making the task impossible to establish.
\item \textbf{1-2:} Invalid judging task. Information is chaotic, completely failing to constitute a valid judging scenario.
\end{itemize}

\subsection{Teaching Material Generation Task Quality}
\paragraph{Description} How good is this Query as an instruction for teaching material generation? The core lies in whether the instruction is clear, specific, and has a clear teaching objective and educational value.

\begin{itemize}
\item \textbf{9-10:} High-quality instruction. The instruction is very specific (e.g., target grade, knowledge point, material type, difficulty), the teaching objective is clear, and it has high practical value.
\item \textbf{7-8:} Good instruction. The instruction is clear, with a definite teaching theme and material type, but might lack some detailed constraints (e.g., difficulty level).
\item \textbf{5-6:} Average instruction. The instruction is too broad, lacks a specific target, but its basic intent can still be understood.
\item \textbf{3-4:} Low-quality instruction. The instruction is vague, or the requested material scope is too large/impractical (e.g., generate a textbook), making it difficult to execute.
\item \textbf{1-2:} Invalid instruction. The instruction is meaningless or completely unrelated to teaching material generation.
\end{itemize}

\subsection{Question Generation Task Quality}
\paragraph{Description} How good is this Query as a generate questions from a knowledge point task? The core lies in whether the provided knowledge point is authentic, specific, and suitable for being transformed into one or more test questions.

\begin{itemize}
\item \textbf{9-10:} High-quality task. The knowledge point is core, specific, highly testable, and comes with clear constraints (e.g., question type, difficulty).
\item \textbf{7-8:} Good task. The knowledge point is valid and testable, but might be somewhat broad, too simple, or lacks specific constraints for question generation.
\item \textbf{5-6:} Usable task. The knowledge point is too trivial or too grand, making question generation challenging, but it's still possible to generate related questions.
\item \textbf{3-4:} Low-quality task. The provided 'knowledge point' is incorrect, vague, or simply unsuitable for generating questions.
\item \textbf{1-2:} Invalid task. No valid knowledge point information is provided.
\end{itemize}

\subsection{Emotional Support Task Quality}
\paragraph{Description} How good is this Query as a scenario for student mental health? The evaluation core is: Is the scenario description typical, authentic, and specific, thus constituting a valuable case for model learning and response?

\begin{itemize}
\item \textbf{9-10:} High-quality case. The scenario description is extremely authentic, specific, and highly representative, accurately capturing the complex emotions and struggles of a student in a specific situation (e.g., detailed description of self-doubt triggered by an exam failure, worry about disappointing parents, and confusion about future studies), forming a high-quality, in-depth typical case.
\item \textbf{7-8:} Good case. The scenario description is authentic and typical, clearly reflecting a common struggle among the broader student population (e.g., social anxiety in a new environment, pressure from excessive homework), making it a good, effective case.
\item \textbf{5-6:} Usable case. The core struggle is real, but the description is too general, simplified, or lacks necessary background information (e.g., merely 'I feel bad, don't want to go to school'), resulting in insufficient targeting of the case.
\item \textbf{3-4:} Low-quality case. The scenario description is unrealistic, exaggerated, sensational, or deliberately creates dramatic conflict, not resembling situations real students encounter, offering low value as a training sample.
\item \textbf{1-2:} Invalid case. The scenario is entirely fictional, logically chaotic, or the content is completely unrelated to the theme of student mental health, unable to constitute a valid task.
\end{itemize}

\subsection{Personalized Content Creation Task Quality}
\paragraph{Description} How good is this Query as a provide advice based on profile task? The core lies in whether the student profile is informative, logically self-consistent, and whether the goal of the advice is clear.

\begin{itemize}
\item \textbf{9-10:} High-quality profile. The profile is rich in multi-dimensional information (including strengths/weaknesses, interests, habits), logically self-consistent, and the goal for seeking advice is very clear.
\item \textbf{7-8:} Good profile. The profile contains multiple valid information points, is logically clear, and the advice goal is clear enough to generate personalized advice.
\item \textbf{5-6:} Usable profile. The profile information is scarce or relatively singular, but can still be used to generate some general advice. The advice goal is not clear enough.
\item \textbf{3-4:} Low-quality profile. The profile information is insufficient, self-contradictory, or completely lacks a direction for the sought advice.
\item \textbf{1-2:} Invalid profile. The profile information is empty or meaningless, cannot serve as a basis for advice.
\end{itemize}

\subsection{Personalized Learning Support Task Quality}
\paragraph{Description} How good is this Query as a design learning path task? The core lies in whether the profile is effective, and whether it provides a clear learning objective (starting point and desired endpoint).

\begin{itemize}
\item \textbf{9-10:} High-quality task. Contains an information-rich student profile and clearly defines the learning starting point, endpoint, and specific goals, possessing high planning value.
\item \textbf{7-8:} Good task. Contains an effective student profile and clear learning objectives, sufficient for planning a reasonable path.
\item \textbf{5-6:} Usable task. Either the profile information or the learning objective is not clear enough, but a general learning path can still be inferred and designed.
\item \textbf{3-4:} Low-quality task. The profile information is poor, and the learning objective is completely undefined, causing the path design to lack basis and direction.
\item \textbf{1-2:} Invalid task. The profile or target information is missing, unable to constitute a valid learning path design task.
\end{itemize}

\section{Evaluation Metric Design Details}
\label{appendix:evaluation_metric_design}
\subsection{Scenario Adaptation Criteria}
\label{appendix:scenario-adaptation}

The Scenario Adaptation metric evaluates whether the model’s output aligns with the scenario-specific expectations and pedagogical goals. Below are detailed descriptions of its four sub-components:

\begin{itemize}
    \item \textbf{Instruction Following \& Task Completion:} This sub-metric measures the model’s ability to accurately interpret and complete assigned tasks, such as solving problems, correcting errors, or generating questions, while adhering to the required output format and constraints.
    
    \item \textbf{Role \& Tone Consistency:} This dimension evaluates whether the language style, tone, and level of expertise in the response are appropriate for the designated role (e.g., teacher, teaching assistant, peer) and the target learner group (e.g., primary school students, university students).
    
    \item \textbf{Content Relevance \& Scope Control:} The response is assessed for its focus on the specified topic or knowledge area, as well as its ability to stay within the intended difficulty level, subject boundaries, and content scope.
    
    \item \textbf{Scenario Element Integration:} This sub-metric measures the degree to which the model effectively incorporates scenario-specific information, such as prior student responses, individual learning preferences, or stated pedagogical objectives. This is especially important in personalized learning and interactive tutoring contexts.
\end{itemize}

\subsection{Factual \& Reasoning Accuracy Criteria}
\label{appendix:factual-reasoning-accuracy}

This metric evaluates whether a model’s response is grounded in factual correctness and logical rigor, particularly in scenario-intensive or multi-step tasks. It includes the following sub-components:

\begin{itemize}
    \item \textbf{Basic Factual Accuracy:} This sub-metric examines the accuracy of objective information, including definitions, formulas, factual statements, code syntax, and terminology.

    \item \textbf{Domain Knowledge Accuracy:} It assesses the appropriateness and depth of subject-specific knowledge presented in the response, ensuring alignment with disciplinary standards across domains such as mathematics, law, and computer science.

    \item \textbf{Reasoning Process Rigor:} This criterion focuses on the completeness and logical validity of the model's reasoning in tasks that require multi-step derivations, explanations, or justifications.

    \item \textbf{Error Identification \& Correction Precision:} In contexts involving diagnostics or feedback, this sub-metric evaluates the model’s ability to accurately detect, localize, and correct errors without introducing false positives or negatives.
\end{itemize}

\subsection{Pedagogical Application Criteria}
\label{appendix:pedagogical-application}

This metric evaluates whether a model's response demonstrates pedagogical effectiveness and contributes meaningfully to learning outcomes. It includes the following sub-components:

\begin{itemize}
    \item \textbf{Clarity, Simplicity \& Inspiration:} This sub-metric assesses whether the explanation is articulated clearly and accessibly, using appropriate language to promote understanding and stimulate student interest or engagement.

    \item \textbf{Motivation, Guidance \& Positive Feedback:} It evaluates the model’s ability to encourage learners through constructive feedback and supportive guidance, promoting confidence and independent thinking rather than relying on direct answers alone.

    \item \textbf{Personalization, Adaptation \& Learning Support:} This criterion measures the response’s ability to adapt based on the learner's background, proficiency level, and individual needs, including tailored suggestions, scaffolded prompts, and relevant resource recommendations.

    \item \textbf{Higher-Order Thinking \& Skill Development:} This sub-metric examines whether the response promotes advanced cognitive skills, such as critical thinking, problem-solving, creative reasoning, and the ability to transfer knowledge to new contexts.
\end{itemize}

\subsection{Metric allocation for each scenario}
\label{appendix:metric-allocation}

To ensure that evaluation is both fair and context-sensitive, we dynamically allocate evaluation metrics based on the instructional characteristics and goals of each scenario. Not all metrics are applicable across all scenarios: for example, reasoning rigor is essential in problem-solving, while emotional and adaptive support is critical in student guidance tasks. The following table summarizes the allocation of the designed metrics (Section~\ref{sec:metric_allocation}) across the nine scenarios in EduBench:

\begin{table*}[h]
\centering
\begin{adjustbox}{width=\textwidth}
\begin{tabular}{l|ccccccccc}
\toprule
\textbf{Evaluation Metric} & 
\fbox{Error Correction} & 
\fbox{Idea Provision} & 
\fbox{Grading} & 
\fbox{Answering Questions} & 
\fbox{Material Generation} & 
\fbox{Question Generation} & 
\fbox{Mental Health} & 
\fbox{Personalized Content Creation } & 
\fbox{Learning Support} \\
\midrule
\textbf{Instruction Following \& Task Completion} & \cmark & \cmark & \cmark & \cmark & \cmark & \cmark & \cmark & \cmark & \cmark \\
\textbf{Role \& Tone Consistency} &  &  &  &  & \cmark &  &\cmark  &  &  \\
\textbf{Content Relevance \& Scope Control} &  & \cmark & \cmark & \cmark & \cmark & \cmark &  &  & \cmark \\
\textbf{Scenario Element Integration} & \cmark & \cmark &  &  &  &  & \cmark & \cmark & \cmark \\
\midrule
\textbf{Basic Factual Accuracy} & \cmark & \cmark & \cmark & \cmark & \cmark & \cmark &  &  &  \\
\textbf{Domain Knowledge Accuracy} &  & \cmark &  &  & \cmark & \cmark &  &  &  \\
\textbf{Reasoning Process Rigor} & \cmark & \cmark & \cmark & \cmark &  &  &  &  &  \\
\textbf{Error Identification \& Correction Precision} & \cmark &  & \cmark &  &  &  &  &  &  \\
\midrule
\textbf{Clarity, Simplicity \& Inspiration} & \cmark & \cmark &  &  & \cmark & \cmark &  &  &  \\
\textbf{Motivation, Guidance \& Positive Feedback} & \cmark &  & \cmark &  &  &  &  \cmark &  &  \\
\textbf{Personalization, Adaptation \& Learning Support} &  &  &  &  &  &  & \cmark & \cmark & \cmark \\
\textbf{Higher-Order Thinking \& Skill Development} &  & \cmark &  &  & \cmark & \cmark &  &  & \cmark \\
\bottomrule
\end{tabular}
\end{adjustbox}
\caption{Allocation of evaluation metrics across nine educational scenarios in EduBench.}
\label{tab:metric-allocation}
\end{table*}

This allocation ensures that each scenario is evaluated according to the dimensions most critical to its pedagogical purpose. For instance, cognitive rigor is emphasized in analytical tasks (e.g., problem-solving and grading), while adaptive support and contextual integration are prioritized in student-facing personalization tasks (e.g., learning path design or mental health feedback). This scenario-aware evaluation design enhances the interpretability, accuracy, and instructional relevance of the benchmark results.

\section{Evaluation Prompt Design}
\label{sec:eval-prompt}

To ensure consistent and fair human evaluation across multiple educational tasks and languages, we carefully design a suite of evaluation prompts aligned with our 12 fine-grained metrics (see Section~\ref{sec:metric-design}). Each prompt is tailored to elicit targeted human judgment on a specific aspect of model behavior, such as factual accuracy, instruction following, reasoning rigor, or pedagogical impact.

\begin{tcolorbox}[
    colback=blue!5!white,
    colframe=blue!75!black,
    title=Model Evaluation Prompt Design,
    fonttitle=\bfseries,
    boxrule=0.5mm,
    arc=2mm,
    left=2mm, right=2mm, top=2mm, bottom=2mm,
    breakable
]
I will provide you with an educational question and its corresponding answer. Please evaluate the given answer based on the provided assessment criteria and scoring principles, and output the score along with the reasons in JSON format.

Scoring principles:
\{principle\}

Question:
\{question\}

Answer:
\{response\}

The JSON format is defined as follows:

\begin{verbatim}
{
  "detailed_scores": [
    {
      "principle": "principle 1",
      "score": 0,
      "reason": ""
    },
    ...
  ]
}
\end{verbatim}
\end{tcolorbox}

\section{Prompt Design Examples}
\label{appendix:prompt-examples}

This appendix provides illustrative examples of the prompt templates used during the question data generation process described in Section~\ref{sec:question-gen}. Each example corresponds to a specific task scenario, domain, and target competency, and is designed to ensure the quality, diversity, and controllability of the generated data. The question design and repsonse generaion prompts are as follows: Problem Solving~\ref{prompt:Problem Solving}, Error Correction~\ref{prompt:Error Correction}, Idea Provision~\ref{prompt:Idea Provision}, Personalized Learning Support~\ref{prompt:Personalized Learning Support}, Emotional Support~\ref{prompt:Emotional Support}, Question Generation~\ref{prompt:Question Generation}, Automatic Grading~\ref{prompt:Automatic Grading}, Teaching Material Generation~\ref{prompt:Teaching Material Generation}, Personalized Content Creation~\ref{prompt:Personalized Content Creation}. 

\subsection{Problem Solving}
\label{prompt:Problem Solving}
\begin{tcolorbox}[
    colback=blue!5!white, 
    colframe=blue!75!black, 
    title=Problem Solving Prompt Design, 
    fonttitle=\bfseries, 
    boxrule=0.5mm, 
    arc=2mm, 
    left=2mm, 
    right=2mm, 
    top=2mm, 
    bottom=2mm 
]

Please freely generate an appropriate question based on the following subject and difficulty level, and provide a standard answer. The question type is \{question\_type\}. \\
If the type is a short-answer question, for certain subjects, provide code and mathematical calculations if necessary. Do not return extra content. \\

Subject: \{subject\} \\
Difficulty Level: \{level\} \\

Return in JSON format:
\begin{verbatim}
"question":
"answer":
\end{verbatim}

\vspace{1em}

\textbf{Description:} This prompt corresponds to the following capability requirement:

- \textbf{Capability Requirement:} Problem-Solving Ability
    
- \textbf{Basic Description:} The ability to solve problems raised by students.
    
- \textbf{Scenario Design:} Answering questions across different subjects.
\end{tcolorbox}
\begin{tcolorbox}[
    colback=blue!5!white, 
    colframe=blue!75!black, 
    title=Prompt Design for Obtaining Model Responses in This Scenario, 
    fonttitle=\bfseries, 
    boxrule=0.5mm, 
    arc=2mm, 
    left=2mm, 
    right=2mm, 
    top=2mm, 
    bottom=2mm, 
    breakable
]
\{question\}. Please generate the corresponding answer based on the question. \\
"Answer": \\
Return in JSON format.
\end{tcolorbox}

\subsection{Error Correction}
\label{prompt:Error Correction}
\begin{tcolorbox}[
    colback=green!5!white, 
    colframe=green!75!black, 
    title=Error Correction Prompt Design, 
    fonttitle=\bfseries, 
    boxrule=0.5mm, 
    arc=2mm, 
    left=2mm, 
    right=2mm, 
    top=2mm, 
    bottom=2mm, 
    breakable
]

You are an expert teacher in all subjects, helping students correct errors. Please freely generate a question and an incorrect student answer based on the following subject and difficulty level, and provide corrections. The question type is \{question\_type\}. \\
If the type is a short-answer question, for certain subjects, provide code and mathematical calculations if necessary. Do not return extra content. \\

Subject: \{subject\} \\
Difficulty Level: \{level\} \\

Return in JSON format:
\begin{verbatim}
"question":
"original_answer":
"corrected_answer":
"correction_explanation":
\end{verbatim}

\vspace{1em}

\textbf{Description:} This prompt corresponds to the following capability requirement:

- \textbf{Capability Requirement:} Error Correction Ability
    
- \textbf{Basic Description:} The ability to identify and correct errors in students' homework, exams, or daily learning, providing targeted improvement suggestions.
    
- \textbf{Scenario Design:} Basic error correction: Input incorrect solutions, identify errors, and correct them (e.g., code, math).
\end{tcolorbox}
\begin{tcolorbox}[
    colback=green!5!white, 
    colframe=green!75!black, 
    title=Prompt Design for Obtaining Model Responses in This Scenario, 
    fonttitle=\bfseries, 
    boxrule=0.5mm, 
    arc=2mm, 
    left=2mm, 
    right=2mm, 
    top=2mm, 
    bottom=2mm, 
    breakable
]
\{question\}\{original\_answer\} You are providing error correction services for students' answers. Please provide a "Corrected Answer" and "Error Explanation" based on this question and the original answer. \\
"Corrected Answer": \\
"Error Explanation": \\
Return in JSON format.
\end{tcolorbox}

\subsection{Idea Provision}
\label{prompt:Idea Provision}
\begin{tcolorbox}[
    colback=orange!5!white, 
    colframe=orange!75!black, 
    title=Idea Provision Prompt Design, 
    fonttitle=\bfseries, 
    boxrule=0.5mm, 
    arc=2mm, 
    left=2mm, 
    right=2mm, 
    top=2mm, 
    bottom=2mm, 
    breakable
]

You are an expert teacher in all subjects, helping students with problem-solving guidance instead of providing standard answers. Please freely generate a question based on the following subject and difficulty level, and provide guidance without giving the answer. The question type is \{question\_type\}. \\
If the type is a short-answer question, for certain subjects, provide code and mathematical calculations if necessary. Do not return extra content. \\

Subject: \{subject\} \\
Difficulty Level: \{level\} \\

Return in JSON format:
\begin{verbatim}
"question":
"provided_guidance":
\end{verbatim}

\vspace{1em}

\textbf{Description:} This prompt corresponds to the following capability requirement:

- \textbf{:} Q\&A Ability
    
- \textbf{Basic Description:} The ability to answer students' learning questions in real-time, covering knowledge point explanations, homework assistance, exam preparation guidance, etc.
    
- \textbf{Scenario Design:}
    - Basic knowledge questions: Explanation of knowledge points (accuracy, simplicity, inspiration).
    - Question-solving analysis: Explanation of problem-solving steps, decomposition of complex problems, analysis of involved knowledge points, and summarization of experience.
    - General: Including study techniques, exam strategies, etc.
\end{tcolorbox}
\begin{tcolorbox}[
    colback=orange!5!white, 
    colframe=orange!75!black, 
    title=Prompt Design for Obtaining Model Responses in This Scenario, 
    fonttitle=\bfseries, 
    boxrule=0.5mm, 
    arc=2mm, 
    left=2mm, 
    right=2mm, 
    top=2mm, 
    bottom=2mm, 
    breakable
]
\{question\}Please provide an approach based on this question.\\ "Provided Approach": \\
Return in JSON format.
\end{tcolorbox}

\subsection{Personalized Learning Support}
\label{prompt:Personalized Learning Support}
\begin{tcolorbox}[
    colback=purple!5!white, 
    colframe=purple!75!black, 
    title=Personalized Learning Support Prompt Design, 
    fonttitle=\bfseries, 
    boxrule=0.5mm, 
    arc=2mm, 
    left=2mm, 
    right=2mm, 
    top=2mm, 
    bottom=2mm, 
    breakable
]

You are a personalized service customization expert, providing tailored services to improve learning efficiency. Please freely generate a specific and appropriate student profile based on the following subject and difficulty level, and provide learning path planning suggestions and personalized recommendations. The question type is \{question\_type\}. \\
If the type is a short-answer question, for certain subjects, provide code and mathematical calculations if necessary. Do not return extra content. \\

Subject: \{subject\} \\
Difficulty Level: \{level\} \\

Return in JSON format:
\begin{verbatim}
"student_profile":
"learning_path_planning_suggestions":
"personalized_recommendations":
\end{verbatim}

\vspace{1em}

\textbf{Description:} This prompt corresponds to the following capability requirement:

- \textbf{Capability Requirement:} Personalized Service
    
- \textbf{Basic Description:} Based on student profiles, provide customized services to enhance learning efficiency.
    
- \textbf{Scenario Design:}
    - Learning path planning: Arrange future courses based on current ability levels and learning goals.
    - Personalized recommendations: Generate or recommend practice exercises and reading materials based on weak knowledge areas and learning habits.
\end{tcolorbox}
\begin{tcolorbox}[
    colback=purple!5!white, 
    colframe=purple!75!black, 
    title=Prompt Design for Obtaining Model Responses in This Scenario, 
    fonttitle=\bfseries, 
    boxrule=0.5mm, 
    arc=2mm, 
    left=2mm, 
    right=2mm, 
    top=2mm, 
    bottom=2mm, 
    breakable
]
\{student\_profile\}
Based on the student profile, provide "Learning Path Planning" and "Personalized Recommendations", \\
"Learning Path Planning": \\
"Personalized Recommendations": \\
Returned in JSON format.
\end{tcolorbox}

\subsection{Emotional Support}
\label{prompt:Emotional Support}
\begin{tcolorbox}[
    colback=red!5!white, 
    colframe=red!75!black, 
    title=Emotional Support Prompt Design, 
    fonttitle=\bfseries, 
    boxrule=0.5mm, 
    arc=2mm, 
    left=2mm, 
    right=2mm, 
    top=2mm, 
    bottom=2mm, 
    breakable
]

You are an intelligent assistant capable of identifying students' emotional states, analyzing the causes of emotional issues, and providing relevant comfort and advice. Please freely generate a multi-turn conversation with a student studying \{subject\}, identify their emotional state, analyze the causes of emotional issues, and provide relevant comfort and advice. The anxiety level is \{anxiety\_level\}. \\
Do not return extra content. \\

Academic Level: \{level\} \\

Return in JSON format:
\begin{verbatim}
"conversation_with_student":
"emotional_state_analysis":
"comfort_and_advice":
\end{verbatim}

\vspace{1em}

\textbf{Description:} This prompt corresponds to the following capability requirement:

- \textbf{Capability Requirement:} Psychological Support
    
- \textbf{Basic Description:} The ability to identify students' emotional states, analyze the causes of emotional issues, and provide relevant comfort and advice.
    
- \textbf{Scenario Design:}
    - Emotional recognition: Identify emotional states (e.g., normal, mild anxiety, severe anxiety) based on conversations with students.
    - Comfort and advice: Provide targeted comfort and advice for different emotional issues (e.g., pre-exam anxiety, post-exam frustration, social isolation).
    - Task scenario: Input: Student's text description or questions; Output: Emotional state recognition, targeted advice, and possible follow-up action suggestions.
    - Evaluation metrics: Recognition accuracy, emotional granularity, cause identification, response relevance, and practicality of advice.

\end{tcolorbox}
\begin{tcolorbox}[
    colback=red!5!white, 
    colframe=red!75!black, 
    title=Prompt Design for Obtaining Model Responses in This Scenario, 
    fonttitle=\bfseries, 
    boxrule=0.5mm, 
    arc=2mm, 
    left=2mm, 
    right=2mm, 
    top=2mm, 
    bottom=2mm, 
    breakable
]
\{conversation\_with\_student\}\{anxiety\_level\}
Please provide "Emotional State Analysis" and "Comfort \& Suggestions" based on the student's emotional state and conversation. \\
"Emotional State Analysis": \\
"Comfort \& Suggestions": \\
Return in JSON format.
\end{tcolorbox}

\subsection{Question Generation}
\label{prompt:Question Generation}
\begin{tcolorbox}[
    colback=yellow!5!white, 
    colframe=yellow!75!black, 
    title=Question Generation Prompt Design, 
    fonttitle=\bfseries, 
    boxrule=0.5mm, 
    arc=2mm, 
    left=2mm, 
    right=2mm, 
    top=2mm, 
    bottom=2mm, 
    breakable
]

You are an expert teacher in all subjects, generating appropriate questions based on knowledge scope and question types. Please freely generate a knowledge point and corresponding question based on the following subject and difficulty level, provide guidance, and give an answer. The question type is \{question\_type\}. \\
If the type is a short-answer question, for certain subjects, provide code and mathematical calculations if necessary. Do not return extra content. \\

Subject: \{subject\} \\
Difficulty Level: \{level\} \\

Return in JSON format:
\begin{verbatim}
"knowledge_point":
"question":
"provided_guidance":
"answer":
\end{verbatim}

\vspace{1em}

\textbf{Description:} This prompt corresponds to the following capability requirement:

- \textbf{Capability Requirement:} Question Generation
    
- \textbf{Basic Description:} Generate questions based on knowledge scope and question types, considering difficulty levels.
    
- \textbf{Scenario Design:}
    - Basic question generation: Generate questions for different difficulty levels, question types, and knowledge scopes.
    - Comprehensive question generation: Cross-reference multiple knowledge points to generate questions.
    - Additional requirements:
        - Provide solutions and step-by-step scoring references for questions.
        - Compile quizzes or exams based on syllabi to form structured classroom assessments.

\end{tcolorbox}
\begin{tcolorbox}[
    colback=yellow!5!white, 
    colframe=yellow!75!black, 
    title=Prompt Design for Obtaining Model Responses in This Scenario, 
    fonttitle=\bfseries, 
    boxrule=0.5mm, 
    arc=2mm, 
    left=2mm, 
    right=2mm, 
    top=2mm, 
    bottom=2mm, 
    breakable
]
\{knowledge\_point\}\{subject\}\{question\_type\}\{level\}

Please generate a question based on the subject, academic level, knowledge point, and question type. \\
"Question": \\
Return in JSON format.

\end{tcolorbox}

\subsection{Automatic Grading}
\label{prompt:Automatic Grading}
\begin{tcolorbox}[
    colback=teal!5!white, 
    colframe=teal!75!black, 
    title=Automatic Grading Prompt Design, 
    fonttitle=\bfseries, 
    boxrule=0.5mm, 
    arc=2mm, 
    left=2mm, 
    right=2mm, 
    top=2mm, 
    bottom=2mm, 
    breakable
]

You need to implement:
1. Objective grading: Grade multiple-choice, true/false, and fill-in-the-blank questions; provide step-by-step scoring for open-ended questions.
2. Subjective grading: Evaluate large assignments and lab reports comprehensively (e.g., workload, completeness, knowledge application).
3. Personalized feedback: Generate constructive feedback, including potential knowledge gaps and learning suggestions.

Please freely generate a question and a student's answer based on the following subject and difficulty level, and grade the answer. The question type is \{question\_type\}. \\
If the type is a short-answer question, for certain subjects, provide code and mathematical calculations if necessary. Do not return extra content. \\

Subject: \{subject\} \\
Difficulty Level: \{level\} \\

Return in JSON format:
\begin{verbatim}
"question":
"student_answer":
"grading":
"grading_details":
"personalized_feedback":
\end{verbatim}

\vspace{1em}

\textbf{Description:} This prompt corresponds to the following capability requirement:

- \textbf{Capability Requirement:} Automated Homework Grading
    
- \textbf{Basic Description:} Automatically grade students' homework and analyze results to provide suggestions.
    
- \textbf{Scenario Design:}
    - Objective grading: Grade multiple-choice, true/false, and fill-in-the-blank questions; provide step-by-step scoring for open-ended questions.
    - Subjective grading: Evaluate large assignments and lab reports comprehensively (e.g., workload, completeness, knowledge application).
    - Personalized feedback: Generate constructive feedback, including potential knowledge gaps and learning suggestions.

\end{tcolorbox}
\begin{tcolorbox}[
    colback=teal!5!white, 
    colframe=teal!75!black, 
    title=Prompt Design for Obtaining Model Responses in This Scenario, 
    fonttitle=\bfseries, 
    boxrule=0.5mm, 
    arc=2mm, 
    left=2mm, 
    right=2mm, 
    top=2mm, 
    bottom=2mm, 
    breakable
]
\{question\}\{student\_answer\}
Please provide "Score", "Scoring Details", and "Personalized Feedback" based on the question and student's answer. \\
"Score": \\
"Scoring Details": \\
"Personalized Feedback": \\
Return in JSON format.
\end{tcolorbox}

\subsection{Teaching Material Generation}
\label{prompt:Teaching Material Generation}
\begin{tcolorbox}[
    colback=cyan!5!white, 
    colframe=cyan!75!black, 
    title=Teaching Material Generation Prompt Design, 
    fonttitle=\bfseries, 
    boxrule=0.5mm, 
    arc=2mm, 
    left=2mm, 
    right=2mm, 
    top=2mm, 
    bottom=2mm, 
    breakable
]

You are responsible for helping teachers generate high-quality teaching materials, including lesson plans, presentations, and lecture notes. Based on textbook chapters or knowledge points, automatically generate structured lesson plans, including learning objectives, key points, difficult points, and classroom activity designs. The question type is \{question\_type\}. \\
If the type is a short-answer question, for certain subjects, provide code and mathematical calculations if necessary. Do not return extra content. \\

Subject: \{subject\} \\
Difficulty Level: \{level\} \\

Return in JSON format:
\begin{verbatim}
"knowledge_point":
"teaching_materials":
\end{verbatim}

\vspace{1em}

\textbf{Description:} This prompt corresponds to the following capability requirement:

- \textbf{Capability Requirement:} Teaching Material Generation
    
- \textbf{Basic Description:} Generate high-quality teaching materials, including lesson plans, presentations, and lecture notes.
    
- \textbf{Scenario Design:}
    - Course PPT or lecture note generation: Automatically generate slides or detailed notes based on textbook chapters or knowledge points.
    - Lesson plan generation: Automatically generate structured lesson plans, including learning objectives, key points, difficult points, and classroom activity designs.
    - Course-related material generation or retrieval: Generate or search for relevant materials, such as images, teaching cases, links, and references.

\end{tcolorbox}
\begin{tcolorbox}[
    colback=cyan!5!white, 
    colframe=cyan!75!black, 
    title=Prompt Design for Obtaining Model Responses in This Scenario, 
    fonttitle=\bfseries, 
    boxrule=0.5mm, 
    arc=2mm, 
    left=2mm, 
    right=2mm, 
    top=2mm, 
    bottom=2mm, 
    breakable
]
\{knowledge\_point\}
Please provide "Teaching Material" based on this knowledge point. The teaching material should include teaching objectives, key points and difficulties, classroom activity design, etc. \\
"Teaching Material": \\
Return in JSON format.
\end{tcolorbox}

\subsection{Personalized Content Creation}
\label{prompt:Personalized Content Creation}
\begin{tcolorbox}[
    colback=purple!5!white, 
    colframe=purple!75!black, 
    title=Personalized Content Creation Prompt Design, 
    fonttitle=\bfseries, 
    boxrule=0.5mm, 
    arc=2mm, 
    left=2mm, 
    right=2mm, 
    top=2mm, 
    bottom=2mm, 
    breakable
]

You are an intelligent assistant capable of generating personalized learning content or tasks based on individual student differences. Please freely generate a student profile for a student studying \{subject\}, and consider the following three aspects:
1. One-on-one: Customize practice questions or reading materials based on specific student profiles.
2. Tiered teaching: For the same course content, generate different teaching objectives, methods, assessment methods, and homework assignments for students at different levels.
3. Other: Combine other capability requirements to design differentiated ability evaluation data for individual students, study groups, and classes.

Do not return extra content.

Academic Level: \{level\}

Strictly return in JSON format:
\begin{verbatim}
"student_profile":
"personalized_learning_content_or_\
tasks":
\end{verbatim}

\vspace{1em}

\textbf{Description:} This prompt corresponds to the following capability requirement:

- \textbf{Capability Requirement:} Personalized Content Generation
    
- \textbf{Basic Description:} Generate personalized learning content or tasks based on individual student differences.
    
- \textbf{Scenario Design:}
    - One-on-one customization: Based on specific student profiles, generate tailored practice questions or reading materials.
    - Tiered teaching: For the same course content, generate different teaching objectives, methods, assessment methods, and homework assignments for students at different levels.
    - Other considerations: Combine other capability requirements to design differentiated ability evaluation data for individual students, study groups, and classes.

\end{tcolorbox}
\begin{tcolorbox}[
    colback=purple!5!white, 
    colframe=purple!75!black, 
    title=Prompt Design for Obtaining Model Responses in This Scenario, 
    fonttitle=\bfseries, 
    boxrule=0.5mm, 
    arc=2mm, 
    left=2mm, 
    right=2mm, 
    top=2mm, 
    bottom=2mm, 
    breakable
]
\{student\_profile\}
Based on the student profile, generate a 'personalized learning content or task' for each student. \\
"personalized learning content or task": \\
Return in JSON format.
\end{tcolorbox}

\section{Comprehensive Evaluation Metric Scoring Details}
\label{app:metric-scores}
To systematically evaluate the quality of AI-generated responses in educational settings, we designed a comprehensive human evaluation rubric comprising three main dimensions, each containing several fine-grained criteria. Each criterion is rated on a 10-point scale with clearly defined level anchors to guide consistent judgment. Details are listed as follows:
\subsection{Instructional Quality}

\subsubsection{Instruction Following \& Task Completion (IFTC)}
\textbf{Description:} Did it fully understand and execute the user's instruction? Was the core task (e.g., solving problems, error correction, question generation) completed? Is the output formatting correct?
\begin{itemize}
    \item \textbf{9-10:} Fully understood and precisely executed all instructions; achieved core task with perfect accuracy; output format is fully compliant.
    \item \textbf{7-8:} Accurately understood main instructions and correctly completed the task; core goals are well achieved; format is mostly correct with only minor omissions or deviations.
    \item \textbf{5-6:} Understood the general intent but may miss some details; task largely completed but with some inaccuracies or omissions; formatting attempts present but with notable flaws.
    \item \textbf{3-4:} Misunderstood part of the instruction; low task completion or major errors; formatting mostly incorrect.
    \item \textbf{1-2:} Completely misunderstood or ignored instructions; task not completed or totally incorrect; formatting is chaotic or irrelevant.
\end{itemize}

\subsubsection{Role \& Tone Consistency (RTC)}
\textbf{Description:} Does the language style, tone, and level of professionalism match the assigned role (e.g., teacher, teaching assistant, peer) and the target learner group (e.g., elementary, college)?

\begin{itemize}
    \item \textbf{9-10:} Excellent role-playing (e.g., teacher/TA); language style, professionalism, and tone (e.g., encouraging/serious) are perfectly aligned with the assumed role and audience.
    \item \textbf{7-8:} Role and tone are mostly consistent and appropriate for the scenario, with minor deviation in individual expressions.
    \item \textbf{5-6:} Attempts to match the role and tone can be seen, but overall consistency is weak; some expressions are disconnected from the role/scenario.
    \item \textbf{3-4:} Significant mismatch in role and tone; comes across as unnatural or inconsistent.
    \item \textbf{1-2:} No reflection of assigned role/tone; expression entirely inconsistent with the scenario.
\end{itemize}

\subsubsection{Content Relevance \& Scope Control (CRSC)}
\textbf{Description:} Is the content tightly aligned with the specified topic, theme, or question? Is it kept within the specified difficulty level, scenario, or scope?

\begin{itemize}
    \item \textbf{9-10:} Content is highly relevant to the specified topic/theme/question; strictly within required difficulty/scope/discipline without redundant or irrelevant information.
    \item \textbf{7-8:} Overall relevance is high; scope control is good with possibly a small amount of slightly off-topic or mildly overreaching information.
    \item \textbf{5-6:} Mostly relevant, but includes some off-topic or out-of-scope content; scope control needs improvement.
    \item \textbf{3-4:} Poor relevance; includes a significant amount of irrelevant information or is largely outside scope.
    \item \textbf{1-2:} Content is largely irrelevant or completely outside the specified scope.
\end{itemize}
\subsubsection{Scenario Element Integration (SEI)}
\textbf{Description:} Did it effectively use scenario-specific information (e.g., previous student answers, learning preferences, specific teaching goals)? Especially important in personalized, Q\&A, or error-correction contexts.

\begin{itemize}
    \item \textbf{9-10:} Fully integrated all key scenario elements (e.g., student history, learning preferences); output is highly personalized and well-matched to the teaching context.
    \item \textbf{7-8:} Used major scenario elements effectively; response is targeted, possibly overlooks minor details but does not affect overall results.
    \item \textbf{5-6:} Some use of scenario information, but integration is shallow; personalization or contextual fit is average.
    \item \textbf{3-4:} Only surface-level reference to scenario information; did not integrate core elements effectively; weak contextual connection.
    \item \textbf{1-2:} Completely ignored scenario-specific information; output is generic, templated, and irrelevant to the scenario.
\end{itemize}

\subsection{Content Accuracy}

\subsubsection{Basic Factual Accuracy (BFA)}
\textbf{Description:} Are objective facts such as concept definitions, formulas, dates, terminology, code syntax, legal clauses correctly presented?

\begin{itemize}
    \item \textbf{9-10:} All stated factual elements (definitions, formulas, dates, terms, syntax, etc.) are completely accurate.
    \item \textbf{7-8:} Vast majority of facts are correct; possibly contains very minor, non-critical typos or omissions.
    \item \textbf{5-6:} Most facts are correct, but there are some notable factual errors that require review.
    \item \textbf{3-4:} Contains several or key factual inaccuracies; information is not trustworthy.
    \item \textbf{1-2:} Riddled with factual errors; information is completely incorrect or misleading.
\end{itemize}

\subsubsection{Domain Knowledge Accuracy (DKA)}
\textbf{Description:} Is the use of subject matter knowledge (math, programming, law, finance, etc.) not only correct but also appropriately specialized and aligned with domain standards?

\begin{itemize}
    \item \textbf{9-10:} Subject matter application is not only accurate but also shows appropriate depth and rigor; adheres to industry or academic standards.
    \item \textbf{7-8:} Proper use of professional knowledge reflecting a good degree of proficiency; minor shortcomings in depth or detail not affecting validity.
    \item \textbf{5-6:} Basic accuracy in subject knowledge, but somewhat surface-level or lacking rigor; some confusion or omissions of non-core concepts.
    \item \textbf{3-4:} Significant errors or major omission in subject-specific content; lacks professionalism.
    \item \textbf{1-2:} Serious domain errors; completely incorrect or misleading; does not meet any professional standards.
\end{itemize}

\subsubsection{Reasoning Process Rigor (RPR)}
\textbf{Description:} For content requiring reasoning (e.g., math steps, code logic, legal arguments, case analysis), is the logical flow complete and sound?

\begin{itemize}
    \item \textbf{9-10:} Reasoning is complete, clear, and rigorous; all steps are correct; arguments are strong and free of logical fallacies.
    \item \textbf{7-8:} Reasoning is largely correct and logically coherent with minor issues in individual steps or details that do not affect the conclusion.
    \item \textbf{5-6:} Reasoning is visible but contains unclear logic, missing steps, or insufficient argumentation, affecting the overall outcome.
    \item \textbf{3-4:} Reasoning has major logical flaws, confusion in steps, or critical omissions; reliability is low.
    \item \textbf{1-2:} Virtually no valid reasoning; logic is chaotic; steps are incorrect or irrelevant.
\end{itemize}

\subsubsection{Error Identification \& Correction Precision (EICP)}
\textbf{Description:} In error correction scenarios, are errors precisely identified (no missed or false positives)? Are the corrections correct and optimal?

\begin{itemize}
    \item \textbf{9-10:} Precisely identified all errors (no omission or false positives); provided completely correct, clear, and optimal correction suggestions.
    \item \textbf{7-8:} Correctly located most major errors; suggestions are generally accurate and effective with only minor omissions or less-than-perfect advice.
    \item \textbf{5-6:} Identified some errors but with clear omissions or false positives; suggestions are partially correct but may lack clarity, completeness or optimality.
    \item \textbf{3-4:} Inaccurate error detection with critical omissions or many false positives; suggestions contain errors or are hard to comprehend.
    \item \textbf{1-2:} Completely failed to detect errors; provided entirely incorrect or misleading correction advice.
\end{itemize}


\subsection{Pedagogical Effectiveness}

\subsubsection{Clarity, Simplicity \& Inspiration (CSI)}
\textbf{Description:} Are explanations, descriptions, and feedback clear, concise, and easy for the target learners to understand? Is the delivery inspiring and thought-provoking?

\begin{itemize}
    \item \textbf{9-10:} Extremely clear and concise explanations; fully accessible for target learners; vibrant and engaging delivery that inspires deep thought and interest.
    \item \textbf{7-8:} Clear and easy to understand; appropriate for learner level; somewhat thought-provoking and can trigger reflection.
    \item \textbf{5-6:} Generally understandable but may be wordy, complex, or dull; limited inspirational impact.
    \item \textbf{3-4:} Lacks clarity; uses excessive jargon or complex structures; difficult to comprehend; uninspiring.
    \item \textbf{1-2:} Confusing and hard to follow; disregards learner needs; offers no inspiration and may cause confusion.
\end{itemize}

\subsubsection{Motivation, Guidance \& Positive Feedback (MGP)}
\textbf{Description:} Does the interaction provide encouragement and support? Is constructive and positive language used? In answering or tutoring, does it guide thinking or just give away answers?

\begin{itemize}
    \item \textbf{9-10:} Strongly supportive and encouraging; consistently uses constructive and positive language; offers highly effective heuristic guidance instead of simply giving answers.
    \item \textbf{7-8:} Generally supportive tone and positive language; provides useful guidance though occasionally too direct.
    \item \textbf{5-6:} A mix of encouragement and neutral/critical language; guidance is inconsistent—sometimes helpful, sometimes overly direct or lacking.
    \item \textbf{3-4:} Lacks encouragement and support; language is neutral or mildly negative; rarely guides, often just answers or remains unhelpful.
    \item \textbf{1-2:} Negative or discouraging tone; no motivation or support; fails to guide or gives misleading suggestions.
\end{itemize}

\subsubsection{Personalization, Adaptation \& Learning Support (PAS)}
\textbf{Description:} Can it provide differentiated content, advice, or feedback based on a student's level, traits, or needs? Does it recommend effective learning paths or resources?

\begin{itemize}
    \item \textbf{9-10:} Highly personalized content/advice/feedback based on student level/traits/needs; resource and learning path suggestions are accurate, practical, and valuable.
    \item \textbf{7-8:} Demonstrates some adaptation to student situation; provides relevant learning advice or resources with good utility.
    \item \textbf{5-6:} Attempts personalization but with limited effectiveness; recommendations are generic and of limited value.
    \item \textbf{3-4:} Little to no personalization; output is the same for everyone; learning support is insufficient or unrelated.
    \item \textbf{1-2:} No personalization; output may conflict with student needs; offers no or incorrect learning support.
\end{itemize}

\subsubsection{Higher-Order Thinking \& Skill Development (HOTS)}
\textbf{Description:} Does the interaction or content help foster students' critical thinking, creativity, problem-solving, or knowledge transfer skills?

\begin{itemize}
    \item \textbf{9-10:} Skillfully designed to promote critical/creative thinking, problem-solving, or transfer of knowledge (e.g., through open-ended questions, comparative analysis, case study, project-based tasks).
    \item \textbf{7-8:} Includes guiding questions or moderately challenging tasks that positively support the development of higher-order thinking (e.g., analysis, evaluation, application).
    \item \textbf{5-6:} Some attempt to encourage higher-order thinking (e.g., simple reflective questions), but limited in depth and scope; mainly focused on rote understanding or basic application.
    \item \textbf{3-4:} Interaction/content mostly revolves around memory and comprehension; rarely addresses higher-order thinking tasks.
    \item \textbf{1-2:} Completely ignores higher-order skill development; encourages rote memorization and repetition; may inhibit thinking flexibility.
\end{itemize}

\section{Human Annotator Cost}
The cost for each QA pair is \$2.22. We provided 198 questions (99 in English, 99 in Chinese) with 5 responses per question, totaling 990 QA pairs. The final cost is approximately \$2,194.

\section{Detailed Correlation Analysis Between Model Evaluation and Human Assessment across Dimensions}
\label{app:detailed_correlation}
\begin{table*}[h]
\centering
\begin{adjustbox}{width=\textwidth}
\begin{tabular}{c|ccccc|ccccc|}
\toprule
\textbf{Model}&\multicolumn{5}{c}{\textbf{Instruction Following \& Task Completion}}&\multicolumn{5}{c}{\textbf{Role \& Tone Consistency}}\\
\midrule
&\textbf{DeepSeek R1} & \textbf{GPT-4o} & \textbf{QwQ-Plus} & \textbf{DeepSeek V3} & \textbf{Human}&\textbf{DeepSeek R1} & \textbf{GPT-4o} & \textbf{QwQ-Plus} & \textbf{DeepSeek V3} & \textbf{Human}\\
\midrule
\textbf{DeepSeek R1} &-&0.56&0.66&0.69&0.63&-&0.48&0.55&0.65&0.71\\
 \textbf{GPT-4o} &0.56&-&0.55&0.52&0.53&0.48&-&0.55&0.56&0.53\\
 \textbf{QwQ-Plus} &0.66&0.55&-&0.63&0.58&0.55&0.55&-&0.58&0.59\\
 \textbf{DeepSeek V3} &0.69&0.52&0.63&-&0.65&0.65&0.56&0.58&-&0.7\\
 \textbf{Human}&0.63&0.53&0.58&\textbf{0.65}&-&\textbf{0.71}&0.53&0.59&0.7&-\\
\midrule
&\multicolumn{5}{c}{\textbf{Content Relevance \& Scope Control}}&\multicolumn{5}{c}{\textbf{Scenario Element Integration}}\\
\midrule
&\textbf{DeepSeek R1} & \textbf{GPT-4o} & \textbf{QwQ-Plus} & \textbf{DeepSeek V3} & \textbf{Human}&\textbf{DeepSeek R1} & \textbf{GPT-4o} & \textbf{QwQ-Plus} & \textbf{DeepSeek V3} & \textbf{Human}\\
\midrule
\textbf{DeepSeek R1} & -&0.57&0.61&0.65&0.61&-&0.55&0.59&0.7&0.66\\
\textbf{GPT-4o} & 0.57&-&0.6&0.6&0.55&0.55&-&0.53&0.56&0.54\\
\textbf{QwQ-Plus} &0.61&0.6&-&0.62&0.64&0.59&0.53&-&0.61&0.63\\
 \textbf{DeepSeek V3} &0.65&0.6&0.62&-&0.61&0.7&0.56&0.61&-&0.7\\
 \textbf{Human}&0.61&0.55&\textbf{0.64}&0.61&-&0.66&0.54&0.63&\textbf{0.7}&-\\
\bottomrule
\end{tabular}
\end{adjustbox}
\caption{\textbf{Kendall’s} $W$ between different evaluation models and human evaluation in Instructional Quality.}
\label{tab:spearman_comparison_IQ}
\end{table*}

\begin{table*}[h]
\centering
\begin{adjustbox}{width=\textwidth}
\begin{tabular}{c|ccccc|ccccc|}
\toprule
\textbf{Model}&\multicolumn{5}{c}{\textbf{Basic Factual Accuracy}}&\multicolumn{5}{c}{\textbf{Domain Knowledge Accurac}}\\
\midrule
&\textbf{DeepSeek R1} & \textbf{GPT-4o} & \textbf{QwQ-Plus} & \textbf{DeepSeek V3} & \textbf{Human}&\textbf{DeepSeek R1} & \textbf{GPT-4o} & \textbf{QwQ-Plus} & \textbf{DeepSeek V3} & \textbf{Human}\\
\midrule
\textbf{DeepSeek R1} &-&0.51&0.66&0.68&0.59&-&0.6&0.59&0.58&0.57\\
 \textbf{GPT-4o} &0.51&-&0.57&0.56&0.59&0.6&-&0.59&0.62&0.56\\
 \textbf{QwQ-Plus} &0.66&0.57&-&0.62&0.63&0.59&0.59&-&0.64&0.64\\
 \textbf{DeepSeek V3} & 0.68&0.56&0.62&-&0.54&0.58&0.62&0.64&-&0.54\\
\textbf{Human}&0.59&0.59&\textbf{0.63}&0.54&-&0.57&0.56&\textbf{0.64}&0.54&-\\
\midrule
&\multicolumn{5}{c}{\textbf{Reasoning Process Rigor}}&\multicolumn{5}{c}{\textbf{Error Identification \& Correction Precision}}\\
\midrule
&\textbf{DeepSeek R1} & \textbf{GPT-4o} & \textbf{QwQ-Plus} & \textbf{DeepSeek V3} & \textbf{Human}&\textbf{DeepSeek R1} & \textbf{GPT-4o} & \textbf{QwQ-Plus} & \textbf{DeepSeek V3} & \textbf{Human}\\
\midrule
\textbf{DeepSeek R1} &-&0.55&0.59&0.69&0.62&-&0.53&0.56&0.54&0.52\\
 \textbf{GPT-4o} &0.55&-&0.57&0.6&0.55&0.53&-&0.55&0.58&0.59\\
 \textbf{QwQ-Plus} &0.59&0.57&-&0.57&0.64&0.56&0.55&-&0.68&0.67\\
 \textbf{DeepSeek V3} & 0.69&0.6&0.57&-&0.65&0.54&0.58&0.68&-&0.66\\
\textbf{Human}&0.62&0.55&0.64&\textbf{0.65}&-&0.52&0.59&\textbf{0.67}&0.66&-\\
\bottomrule
\end{tabular}
\end{adjustbox}
\caption{\textbf{Kendall’s} $W$ between different evaluation models and human evaluation in Content Accuracy.}
\label{tab:spearman_comparison_CA}
\end{table*}

\begin{table*}
\centering
\begin{adjustbox}{width=1\textwidth}
\begin{tabular}{c|ccccc|ccccc|}
\toprule
\textbf{Model}&\multicolumn{5}{c}{\textbf{Clarity, Simplicity \& Inspiration}}&\multicolumn{5}{c}{\textbf{Motivation, Guidance \& Positive Feedback}}\\
\midrule
&\textbf{DeepSeek R1} & \textbf{GPT-4o} & \textbf{QwQ-Plus} & \textbf{DeepSeek V3} & \textbf{Human}&\textbf{DeepSeek R1} & \textbf{GPT-4o} & \textbf{QwQ-Plus} & \textbf{DeepSeek V3} & \textbf{Human}\\
\midrule
\textbf{DeepSeek R1} &-&0.56&0.56&0.54&0.57&-&0.54&0.58&0.59&0.67\\
 \textbf{GPT-4o} &0.56&-&0.57&0.52&0.51&0.54&-&0.58&0.57&0.58\\
 \textbf{QwQ-Plus} &0.56&0.57&-&0.56&0.56&0.58&0.58&-&0.53&0.61\\
 \textbf{DeepSeek V3} &0.54&0.52&0.56&-&0.56&0.59&0.57&0.53&-&0.54\\
 \textbf{Human}&\textbf{0.57}&0.51&0.56&0.56&-&\textbf{0.67}&0.58&0.61&0.54&-\\
\midrule
&\multicolumn{5}{c}{\textbf{Personalization, Adaptation \& Learning Support}}&\multicolumn{5}{c}{\textbf{Higher-Order Thinking \& Skill Development}}\\
\midrule
&\textbf{DeepSeek R1} & \textbf{GPT-4o} & \textbf{QwQ-Plus} & \textbf{DeepSeek V3} & \textbf{Human}&\textbf{DeepSeek R1} & \textbf{GPT-4o} & \textbf{QwQ-Plus} & \textbf{DeepSeek V3} & \textbf{Human}\\
\midrule
\textbf{DeepSeek R1} &-&0.61&0.71&0.74&0.74&-&0.58&0.69&0.69&0.68\\
 \textbf{GPT-4o} & 0.61&-&0.56&0.62&0.59&0.58&-&0.62&0.6&0.6\\
\textbf{QwQ-Plus} &0.71&0.56&-&0.71&0.71&0.69&0.62&-&0.68&0.66\\
 \textbf{DeepSeek V3} &0.74&0.62&0.71&-&0.72&0.69&0.6&0.68&-&0.68\\
 \textbf{Human}&\textbf{0.74}&0.59&0.71&0.72&-&\textbf{0.68}&0.6&0.66&\textbf{0.68}&-\\
\bottomrule
\end{tabular}
\end{adjustbox}
\caption{\textbf{Kendall’s} $W$ between different evaluation models and human evaluation in Pedagogical Effectiveness.}
\label{tab:spearman_comparison_PE}
\end{table*}

\section{Pre-Experiments}

\subsection{Existence of self-preference}
\label{sec:self-pref}
Given the potential influence of both the response generation model and the evaluation model, it is crucial to verify whether a model tends to favor its own responses. To examine this, we use each of the three response-generation models as evaluators to assess the responses they themselves generated. Specifically, for each question, we construct pairwise comparisons among the responses generated by the three models. The evaluator model is then asked to select the better response from each pair. By counting the number of times each model’s output is preferred and comparing the distribution of win rates across different evaluators, we assess whether models exhibit self-preference biases.
\begin{table}[t]
    \centering
    \begin{adjustbox}{width=\linewidth}
    \begin{tabular}{ccc}
        \toprule
        \textbf{Evaluator} & \textbf{Win times} \\
        \midrule
        Qwen Max          & Qwen Max(68) $<$ DeepSeek V3(101) $<$ DeepSeek R1(193) \\
        DeepSeek V3       & Qwen Max(82) $<$ DeepSeek V3(86) $<$ DeepSeek R1(196) \\
        DeepSeek R1       & Qwen Max(68) $<$ DeepSeek V3(110) $<$ DeepSeek R1(186) \\
        \bottomrule
    \end{tabular}
    \end{adjustbox}
    \caption{To investigate whether models exhibit a preference for their own generations, we employ each data-generating models as evaluators to assess the datasets they produced. Specifically, we construct pairwise comparisons by selecting answers generated by two different models at a time. The evaluator is then asked to choose the better response from each pair. By aggregating the number of times each model’s outputs are preferred, we examine potential self-preference biases.}
    \label{tab:self-pref}
\end{table}

As shown in the Table~\ref{tab:self-pref}, the overall trends are consistent across the three different evaluators, with no substantial differences observed in the exact win rates. Notably, none of the evaluators exhibit a strong preference for the responses they themselves generated. These results suggest that model self-preference is not a significant concern, thereby granting us greater flexibility in the choice of evaluation models.

\subsection{Discrimination between normal model and reasoning model}
After ruling out self-preference biases, we further investigate whether models of different types—such as reasoning models and normal models—exhibit biases toward each other. To this end, we design an experiment in which two external models, QwQ (a reasoning model) and GPT-4o (a normal model), are used to evaluate the responses generated by DeepSeek V3 and DeepSeek R1. The evaluation follows the same pairwise comparison protocol as in the self-preference setting: for each question, the evaluator selects the better response from a pair. By comparing the evaluation outcomes across the two evaluators, we aim to assess potential inter-model biases.

\begin{table}[h]
    \centering
    \begin{tabular}{lll}
        \toprule
        \textbf{Evaluator} & \textbf{Gen\_Models} & \textbf{Win\_times} \\
        \midrule
        \multirow{2}{*}{\textbf{Normal}}     & Normal    & 442 \\
                                               & Reasoning & 399 \\
        \midrule
        \multirow{2}{*}{\textbf{Reasoning}}  & Reasoning & 914 \\
                                               & Normal    & 100 \\
        \bottomrule
    \end{tabular}
    \caption{Comparison of evaluation results between different model types.}
    \label{tab:discrimination}
\end{table}

As shown in Table~\ref{tab:discrimination}, the reasoning model exhibits a clear preference for responses generated by other reasoning models, whereas the general-purpose model demonstrates a more balanced evaluation. Notably, the reasoning model’s bias is substantial, with a win ratio as skewed as 9 to 1 in favor of reasoning models. These results highlight the importance of incorporating both reasoning and normal evaluators in the assessment process to mitigate evaluation bias. Relying on only one type of model may lead to unfair or distorted conclusions.

\subsection{Discrimination between different positions}
\label{sec:pos_pref}
When using pairwise evaluation, a natural concern arises as to whether evaluators might exhibit positional bias—i.e., favoring responses based on their order of presentation. To control for this factor, we randomly sample 200 instances from the evaluation dataset and present them to the evaluator in both the original and reversed order. For each comparison, we record the position of the selected response (e.g., incrementing the Former count if the first response is chosen, and Latter if the second is chosen). This allows us to assess whether response position systematically influences evaluation outcomes.

\begin{table}[h]
    \centering
    \begin{adjustbox}{width=1\linewidth}
    \begin{tabular}{llrr}
        \toprule
        \textbf{Evaluator} & \textbf{Order} & \textbf{Former\_win} & \textbf{Latter\_win} \\
        \midrule
        \multirow{2}{*}{\textbf{GPT-4o}} & Normal  & \textbf{873} & 117 \\
                                         & Reverse & \textbf{742} & 248 \\
        \midrule
        \multirow{2}{*}{\textbf{QwQ}}    & Normal  & 605          & 385 \\
                                         & Reverse & 387          & 603 \\
        \bottomrule
    \end{tabular}
    \end{adjustbox}
    \caption{The table presents the evaluation results after reversing the order of the responses, allowing us to examine whether the evaluator exhibits any positional bias.}
    \label{tab:position}
\end{table}

The experimental results reveal that GPT-4o exhibits a notable positional bias, with the former response being selected significantly more often than the latter, even after the order is reversed. As shown in Table~\ref{tab:position}, when the evaluated responses are identical aside from their order, the former response is chosen at a disproportionately high rate—by a factor of 3 to 7—compared to the latter. In contrast, QwQ demonstrates a more balanced evaluation, with selection counts remaining consistent before and after the reversal, indicating minimal positional bias.

\begin{table*}[h]
    \centering
    \begin{adjustbox}{width=1\linewidth}
        \begin{tabular}{l l r r r | r r r}
            \toprule
            \textbf{Scenario} 
            & \textbf{Category Dimensions} 
            & \multicolumn{3}{c}{\textbf{All Data}}  
            & \multicolumn{3}{c}{\textbf{For Training}} \\
            \cmidrule(lr){3-5} \cmidrule(lr){6-8}
            & 
            & \textbf{Chinese} & \textbf{English} 
            & \textbf{Total} 
            & \textbf{Chinese} & \textbf{English} & \textbf{Total} \\
            \midrule
            Problem Solving & Duration*Difficulty*Subject & 1,306 & 1,328 & 2,634 & 1,272 & 1,284 & 2,556 \\
            Error Correction & Duration*Difficulty*Subject & 620 & 1,350 & 1,970 & 603 & 1,334 & 1,937 \\
            Idea Provision & Duration*Difficulty*Subject & 1,342 & 1,350 & 2,692 & 1,300 & 1,322 & 2,622 \\
            Personalized Learning Support & Duration*Subject & 348 & 561 & 909 & 67 & 435 & 502 \\
            Emotional Support & Duration*Anxiety Level & 1,344 & 1,074 & 2,418 & 1,331 & 1,059 & 2,390 \\
            Question Generation & Duration*Difficulty*Subject & 1,358 & 1,338 & 2,696 & 1,331 & 1,322 & 2,653 \\
            Automatic Grading & Duration*Difficulty*Subject & 931 & 1,073 & 2,004 & 912 & 1,058 & 1,970 \\
            Teaching Material Generation & Duration*Difficulty*Subject & 1,324 & 1,347 & 2,671 & 1,306 & 1,255 & 2,561 \\
            Personalized Content Creation & Duration*Subject & 568 & 259 & 827 & 557 & 235 & 792 \\
            \midrule
            \textbf{Total} & & 9,141 & 9,680 & 18,821 & 8,679 & 9,304 & 17,983 \\
            \bottomrule
        \end{tabular}
    \end{adjustbox}
    \caption{The number of scenarios in the dataset and the specific quantities for each scenario.}
    \label{tab:scenario_data}
\end{table*}

\section{Extra Results}
\label{sec:extra-results}
We present the additional experimental results in this section, including the evaluation results of the five response models by the three additional evaluators: GPT-4o, DeepSeek R1, and QwQ-Plus. The evaluation is conducted across two dimensions: metric-level and task-level assessments. The task-level score is the average score of the metrics under each task. The detailed results can be found in Table~\ref{tab:metric-results} and Table~\ref{tab:task-results}.

\begin{table*}[h]
\centering
\begin{adjustbox}{width=0.75\textwidth}
\begin{tabular}{l|l|ccccccccc|c}
\toprule
\textbf{Evaluator} & \textbf{Model} & \textbf{Q\&A} & \textbf{PLS} & \textbf{EC} & \textbf{IP} & \textbf{AG} & \textbf{TMG} & \textbf{ES} & \textbf{QG} & \textbf{PCC} & \textbf{Average} \\
\midrule
\multirow{5}{*}{DeepSeek R1}
& DeepSeek R1            & \textbf{9.81} & \textbf{9.83} & \textbf{9.05} & \textbf{9.11} & 7.74 & \textbf{9.46} & \textbf{9.71} & \textbf{9.22} & \textbf{9.73} & \textbf{9.29} \\
& DeepSeek V3            & 9.67 & 9.12 & 8.97 & 8.82 & \textbf{8.32} & 9.31 & 9.34 & 8.65 & 9.23 & 9.05 \\
& Qwen Max               & 9.07 & 9.11 & 8.86 & 8.84 & 7.99 & 9.15 & 9.40 & 8.89 & 9.29 & 8.96 \\
& Qwen2.5-14B-Instruct   & 8.94 & 8.79 & 8.68 & 8.23 & 7.83 & 9.06 & 8.52 & 8.35 & 8.80 & 8.58 \\
& Qwen2.5-7B-Instruct    & 8.34 & 9.01 & 8.64 & 8.16 & 6.64 & 9.33 & 8.75 & 8.23 & 9.06 & 8.46 \\
\midrule
\multirow{5}{*}{DeepSeek V3}
& DeepSeek R1            & 9.49 & \textbf{9.65} & \textbf{9.27} & \textbf{8.75} & \textbf{7.27} & \textbf{9.45} & \textbf{9.38} & \textbf{9.33} & \textbf{9.71} & \textbf{9.14} \\
& DeepSeek V3            & \textbf{9.68} & 9.04 & 9.14 & 8.53 & 7.05 & 9.34 & 9.00 & 9.06 & 8.92 & 8.86 \\
& Qwen Max               & 9.18 & 8.88 & 9.06 & 8.52 & 7.23 & 9.24 & 9.04 & 9.05 & 9.29 & 8.83 \\
& Qwen2.5-14B-Instruct   & 9.07 & 8.72 & 8.97 & 8.30 & 6.77 & 9.21 & 8.74 & 9.02 & 8.80 & 8.62 \\
& Qwen2.5-7B-Instruct    & 9.15 & 9.07 & 9.01 & 8.47 & 6.44 & 9.21 & 8.85 & 8.69 & 9.00 & 8.65 \\
\midrule
\multirow{5}{*}{GPT-4o}
& DeepSeek R1            & 9.32 & \textbf{9.38} & 9.05 & 8.78 & 8.51 & \textbf{9.25} & \textbf{9.15} & 8.98 & \textbf{9.08} & \textbf{9.06} \\
& DeepSeek V3            & 9.22 & 9.15 & \textbf{9.14} & 8.77 & 8.54 & 9.12 & 9.05 & \textbf{9.00} & 8.95 & 8.99 \\
& Qwen Max               & \textbf{9.50} & 9.17 & 9.01 & 8.69 & \textbf{8.70} & 8.99 & 8.96 & 8.92 & 9.05 & 8.99 \\
& Qwen2.5-14B-Instruct   & 9.34 & 9.25 & 8.92 & 8.51 & 8.11 & 8.99 & 9.11 & 8.77 & 8.82 & 8.87 \\
& Qwen2.5-7B-Instruct    & 9.22 & 9.17 & 8.92 & \textbf{8.84} & 8.04 & 9.05 & 9.00 & 8.62 & 8.94 & 8.87 \\
\midrule
\multirow{5}{*}{QwQ-Plus}
& DeepSeek R1            & \textbf{9.85} & \textbf{9.87} & \textbf{9.24} & \textbf{9.05} & \textbf{8.78} & \textbf{9.75} & \textbf{9.85} & \textbf{9.09} & \textbf{9.88} & \textbf{9.49} \\
& DeepSeek V3            & 9.59 & 9.43 & 9.06 & 8.66 & 8.18 & 9.29 & 9.66 & 8.47 & 9.24 & 9.06 \\
& Qwen Max               & 9.90 & 9.25 & 9.03 & 8.78 & 8.11 & 9.54 & 9.56 & 8.79 & 9.70 & 9.18 \\
& Qwen2.5-14B-Instruct   & 9.83 & 9.21 & 9.05 & 8.23 & 7.88 & 9.22 & 9.45 & 8.48 & 9.02 & 8.94 \\
& Qwen2.5-7B-Instruct    & 9.02 & 9.28 & 8.79 & 8.82 & 7.16 & 9.33 & 9.31 & 7.98 & 9.35 & 8.78 \\
\midrule
\multirow{5}{*}{Human}
& DeepSeek R1            & 7.17 & \textbf{9.11} & \textbf{8.71} & \textbf{8.80} & \textbf{8.42} & \textbf{8.86} & \textbf{9.15} & \textbf{8.79} & \textbf{9.35} & \textbf{8.71} \\
& DeepSeek V3            & 7.45 & 8.12 & 8.16 & 8.17 & 7.84 & 7.56 & 8.08 & 8.01 & 7.03 & 7.82 \\
& Qwen Max               & \textbf{7.72} & 7.94 & 8.21 & 8.15 & 7.89 & 7.99 & 7.85 & 8.39 & 8.42 & 8.06 \\
& Qwen2.5-14B-Instruct   & 7.66 & 7.38 & 7.92 & 7.56 & 7.55 & 7.84 & 7.31 & 7.91 & 7.36 & 7.61 \\
& Qwen2.5-7B-Instruct    & 6.78 & 7.63 & 7.93 & 7.74 & 6.79 & 7.86 & 7.79 & 7.55 & 7.42 & 7.50 \\
\bottomrule
\end{tabular}
\end{adjustbox}
\caption{Scenario-Level Average Scores Evaluated by Different Evaluators. Max values in each column per evaluator are bolded. Full names of each scenarios can be found in Section~\ref{sec:domain_design}.}
\label{tab:task-results}
\end{table*}

\begin{table*}[h]
\centering
\begin{adjustbox}{width=0.85\textwidth}
\begin{tabular}{l|l|cccccccccccc|c}
\toprule
\textbf{Evaluator} & \textbf{Model} & \textbf{BFA} & \textbf{CSI} & \textbf{CRSC} & \textbf{DKA} & \textbf{EICP} & \textbf{HOTS} & \textbf{IFTC} & \textbf{MGP} & \textbf{PAS} & \textbf{RPR} & \textbf{RTC} & \textbf{SEI} & \textbf{Average} \\
\midrule
\multirow{5}{*}{DeepSeek R1} 
& DeepSeek R1            & 9.55 & \textbf{8.67} & \textbf{9.64} & \textbf{9.53} & 8.66 & \textbf{8.39} & \textbf{9.61} & 7.30 & \textbf{9.80} & \textbf{9.17} & \textbf{9.64} & \textbf{9.45} & \textbf{9.12} \\
& DeepSeek V3            & \textbf{9.58} & 8.47 & 9.48 & 9.30 & \textbf{9.32} & 7.53 & 9.39 & \textbf{7.48} & 8.92 & 9.05 & 9.32 & 9.10 & 8.91 \\
& Qwen Max               & 9.42 & 8.49 & 9.46 & 9.24 & 9.09 & 7.67 & 9.25 & 7.44 & 8.97 & 8.62 & 9.34 & 9.05 & 8.84 \\
& Qwen2.5-14B-Instruct   & 9.08 & 8.28 & 9.20 & 8.82 & 8.98 & 7.16 & 8.87 & 6.86 & 8.20 & 8.57 & 9.02 & 8.51 & 8.46 \\
& Qwen2.5-7B-Instruct    & 8.73 & 8.22 & 9.00 & 9.00 & 8.30 & 7.27 & 8.72 & 6.61 & 8.68 & 8.05 & 9.23 & 8.55 & 8.36 \\
\midrule
\multirow{5}{*}{DeepSeek V3} 
& DeepSeek R1            & 9.51 & \textbf{8.75} & \textbf{9.44} & \textbf{9.45} & \textbf{7.61} & \textbf{8.53} & \textbf{9.47} & 7.76 & \textbf{9.64} & \textbf{8.85} & \textbf{9.14} & \textbf{9.06} & \textbf{8.93} \\
& DeepSeek V3            & \textbf{9.57} & 8.61 & 9.25 & 9.27 & 7.23 & 7.98 & 9.21 & 7.56 & 8.94 & 8.76 & 9.00 & 8.59 & 8.66 \\
& Qwen Max               & 9.38 & 8.53 & 9.12 & 9.23 & 7.43 & 7.99 & 9.16 & \textbf{7.85} & 9.05 & 8.57 & 9.00 & 8.61 & 8.66 \\
& Qwen2.5-14B-Instruct   & 9.28 & 8.50 & 9.03 & 9.14 & 7.14 & 7.81 & 8.94 & 7.55 & 8.71 & 8.35 & 8.82 & 8.25 & 8.46 \\
& Qwen2.5-7B-Instruct    & 9.27 & 8.55 & 9.08 & 9.12 & 6.77 & 7.86 & 8.96 & 7.05 & 8.95 & 8.42 & 8.82 & 8.53 & 8.44 \\
\midrule
\multirow{5}{*}{GPT-4o} 
& DeepSeek R1            & 9.48 & \textbf{8.73} & \textbf{9.59} & \textbf{9.17} & 9.05 & \textbf{8.35} & 9.13 & 8.45 & \textbf{9.18} & \textbf{8.89} & \textbf{9.11} & \textbf{8.65} & \textbf{8.98} \\
& DeepSeek V3            & 9.54 & 8.72 & 9.51 & 9.05 & \textbf{9.14} & 8.05 & \textbf{9.16} & \textbf{8.59} & 8.95 & 8.75 & 9.02 & 8.63 & 8.93 \\
& Qwen Max               & \textbf{9.58} & 8.65 & 9.43 & 8.83 & 9.07 & 8.08 & 9.14 & 8.56 & 8.97 & 8.89 & 8.95 & 8.64 & 8.90 \\
& Qwen2.5-14B-Instruct   & 9.45 & 8.51 & 9.44 & 8.88 & 8.93 & 7.83 & 9.02 & 8.20 & 8.88 & 8.60 & 9.07 & 8.43 & 8.77 \\
& Qwen2.5-7B-Instruct    & 9.45 & 8.57 & 9.38 & 8.85 & 8.59 & 8.00 & 9.01 & 8.20 & 8.85 & 8.65 & 9.02 & \textbf{8.65} & 8.77 \\
\midrule
\multirow{5}{*}{QwQ-Plus} 
& DeepSeek R1            & \textbf{9.78} & \textbf{8.47} & \textbf{9.78} & \textbf{9.82} & \textbf{9.70} & \textbf{8.19} & \textbf{9.65} & \textbf{8.35} & \textbf{9.86} & \textbf{9.61} & \textbf{9.70} & \textbf{9.58} & \textbf{9.37} \\
& DeepSeek V3            & 9.42 & 8.25 & 9.57 & 9.09 & 9.52 & 7.22 & 9.36 & 7.62 & 9.23 & 9.23 & 9.39 & 9.32 & 8.93 \\
& Qwen Max               & 9.64 & 8.39 & 9.59 & 9.47 & 9.30 & 7.48 & 9.45 & 7.68 & 9.39 & 9.10 & 9.48 & 9.36 & 9.03 \\
& Qwen2.5-14B-Instruct   & 9.49 & 8.20 & 9.48 & 8.98 & 9.20 & 7.10 & 9.15 & 7.64 & 8.77 & 8.83 & 9.41 & 9.06 & 8.78 \\
& Qwen2.5-7B-Instruct    & 9.08 & 8.10 & 9.31 & 8.98 & 8.91 & 7.02 & 9.03 & 7.18 & 9.09 & 8.61 & 9.30 & 9.33 & 8.66 \\
\midrule
\multirow{5}{*}{Human}
& DeepSeek R1            & \textbf{8.97} & \textbf{8.60} & \textbf{8.98} & \textbf{8.94} & \textbf{8.86} & \textbf{8.56} & \textbf{8.77} & \textbf{8.20} & \textbf{9.26} & \textbf{7.95} & \textbf{8.91} & \textbf{8.92} & \textbf{8.74} \\
& DeepSeek V3            & 8.77 & 7.77 & 8.40 & 7.89 & 8.11 & 7.25 & 8.10 & 7.70 & 7.42 & 7.03 & 7.80 & 7.47 & 7.89 \\
& Qwen Max               & 8.81 & 8.01 & 8.52 & 8.27 & 8.23 & 7.59 & 8.10 & 7.70 & 7.89 & 7.31 & 8.09 & 7.74 & 8.02 \\
& Qwen2.5-14B-Instruct   & 8.74 & 7.76 & 8.26 & 7.79 & 7.86 & 6.88 & 7.77 & 6.97 & 7.02 & 7.01 & 7.59 & 7.03 & 7.56 \\
& Qwen2.5-7B-Instruct    & 8.49 & 7.63 & 8.04 & 7.82 & 7.45 & 6.93 & 7.65 & 7.05 & 7.38 & 5.90 & 7.82 & 7.35 & 7.46 \\
\bottomrule
\end{tabular}
\end{adjustbox}
\caption{Metric-level average scores evaluated by different evaluators under various metrics. For simplicity, we use abbreviations for the metrics. Full names of each metric can be found in Table~\ref{tab:metric-abbreviations}.}
\label{tab:metric-results}
\end{table*}

\section{Distillation Training Setting}
\label{app:data_selection}

After obtaining human and model evaluation results on sample data, we can optimize the selection of model-generated data based on evaluation scores to maximize data quality and efficiency.

We propose two data selection strategies. The first strategy involves selecting the best generation model within each scenario. The specific process includes calculating the average scores of all evaluation metrics from both human and model evaluators on the sample data, ranking the generation models in each scenario according to these average values, and then selecting samples generated by the best-performing model in each scenario from all distilled data.

The second strategy focuses on selecting optimal models for each evaluation metric. We calculate the average score of each generation model on individual metrics, rank them accordingly to identify the best model for each metric. During the final distilled data screening phase, if a piece of data was generated by a model that has been recognized as optimal in any evaluation metric, it will be included in the final finetuning dataset.

We use the following settings for model training: The learning rate is set to $1.0 \times 10^{-5}$, with a batch size of 1 per GPU device. Gradient accumulation is applied over 8 steps, resulting in an effective batch size of 8. For parameter updating, we employ full fine-tuning, where all model parameters are updated during training. All experiments are conducted on 4 NVIDIA A100 GPUs, each with 40GB of memory.

\begin{table*}[t]
\centering
\begin{adjustbox}{width=0.8\textwidth}
\begin{tabular}{l|cccccccccccc|c}
\toprule
\textbf{Model} & \textbf{BFA} & \textbf{CSI} & \textbf{CRSC} & \textbf{DKA} & \textbf{EICP} & \textbf{HOTS} & \textbf{IFTC} & \textbf{MGP} & \textbf{PAS} & \textbf{RPR} & \textbf{RTC} & \textbf{SEI} & \textbf{Average} \\
\midrule
DeepSeek R1            & \textbf{9.21} & \textbf{8.80} & \textbf{9.26} & \textbf{9.41} & \textbf{9.02} & \textbf{8.77} & \textbf{9.10} & \textbf{8.07} & \textbf{9.41} & \textbf{8.13} & 9.25 & \textbf{8.94} & \textbf{8.95} \\
DeepSeek V3            & 9.01 & 7.05 & 8.73 & 8.16 & 8.18 & 6.87 & 8.38 & 7.43 & 7.53 & 7.05 & 9.21 & 7.66 & 7.94 \\
Qwen Max               & 9.05 & 7.41 & 8.85 & 8.55 & 8.21 & 7.54 & 8.52 & 7.53 & 8.24 & 7.41 & 9.25 & 8.05 & 8.22 \\
Qwen2.5-14B-Instruct   & 8.96 & 7.13 & 8.56 & 7.92 & 8.00 & 6.54 & 8.01 & 6.76 & 6.98 & 7.13 & 9.10 & 7.20 & 7.69 \\
Qwen2.5-7B-Instruct    & 8.81 & 6.14 & 8.37 & 8.01 & 7.33 & 6.77 & 7.99 & 6.94 & 7.55 & 8.09 & \textbf{9.26} & 7.60 & 7.74 \\
\bottomrule
\end{tabular}
\end{adjustbox}
\caption{Metric-Level Human Evaluation Results over Three Rounds}
\end{table*}

\begin{table*}[t]
\centering
\begin{adjustbox}{width=0.82\textwidth}
\begin{tabular}{l|ccccccccc|c}
\toprule
\textbf{Model} & \textbf{Q\&A} & \textbf{PLS} & \textbf{EC} & \textbf{IP} & \textbf{AG} & \textbf{TMG} & \textbf{ES} & \textbf{QG} & \textbf{PCC} & \textbf{Average} \\
\midrule
DeepSeek R1            & 7.65 & \textbf{9.56} & \textbf{8.83} & \textbf{9.05} & \textbf{8.40} & \textbf{9.09} & \textbf{9.30} & \textbf{8.91} & \textbf{9.42} & \textbf{8.91} \\
DeepSeek V3            & 7.58 & 7.37 & 8.32 & 8.34 & 7.79 & 8.01 & 8.35 & 8.17 & 7.82 & 7.97 \\
Qwen Max               & \textbf{7.95} & 8.79 & 8.27 & 8.43 & 7.78 & 8.34 & 8.37 & 8.69 & 8.21 & 8.31 \\
Qwen2.5-14B-Instruct   & 7.93 & 7.44 & 7.89 & 7.79 & 7.58 & 7.92 & 7.60 & 8.16 & 7.21 & 7.72 \\
Qwen2.5-7B-Instruct    & 7.04 & 7.64 & 8.07 & 8.16 & 6.77 & 8.21 & 8.12 & 7.73 & 7.53 & 7.70 \\
\bottomrule
\end{tabular}
\end{adjustbox}
\caption{Task-Level Human Evaluation Results over Three Rounds}
\end{table*}

\end{document}